\documentclass[journal, onecolumn]{IEEEtran}
\usepackage{xcolor}
\usepackage{float}
\usepackage{wrapfig}
\usepackage{makecell}
\usepackage{physics}
\usepackage{amsfonts,amssymb,amsmath}
\allowdisplaybreaks
\usepackage{multirow}

\usepackage{amsmath, bm,setspace}
\usepackage{graphicx}
\usepackage{caption}
\usepackage{subcaption}
\usepackage{bbold}
\usepackage[norelsize, linesnumbered, ruled, lined, boxed, commentsnumbered]{algorithm2e}
\usepackage{thmtools,thm-restate}
\usepackage{amsthm}
\usepackage{hyperref}

\newcommand{\meas}{\rho}

\DeclareMathOperator*{\argmin}{argmin}

\newcommand{\ymb}{\mathbf{y}}
\newcommand{\dmb}{\mathbf{d}}

\newcommand{\Qmb}{\mathbf{Q}}
\newcommand{\Ymb}{\mathbf{Y}}

\newcommand{\qmb}{\mathbf{q}}

\newcommand{\Imb}{\mathbf{I}}
\newcommand{\Onemb}{\mathbb{1}}
\newcommand{\realR}{\mathbb{R}}
\newcommand{\T}{N}
\newcommand{\TP}{\mathsf{T}}
\newcommand{\QuantP}{P^{-1}}

\newcommand{\bmb}{\boldsymbol{b}}
\newcommand{\xmb}{\boldsymbol{x}}
\newcommand{\Xmb}{\boldsymbol{X}}

\newif\ifrevision
 \revisionfalse
\makeatletter
\ifrevision
    \usepackage{ulem}
    \def\rev#1{{$\!$\color{red} #1}}
\else 
    \def\rev#1{#1}
\fi

\newtheorem{definition}{Definition}

\newif\ifcomments
 \commentstrue
\makeatletter
\ifcomments
    \usepackage{ulem}
    \def\kcedit#1{{$\!$\color{magenta} [KC: #1]}}
    
\else 
    \def\kcedit#1{}
    
\fi

\ifcomments
    \usepackage{ulem}
    \def\saedit#1{{$\!$\color{blue} [SA: #1]}}
    
\else 
    \def\saedit#1{}
    
\fi

\ifcomments
    \usepackage{ulem}
    \def\elmedit#1{{$\!$\color{red} [ELM: #1]}}
    
\else 
    \def\elmedit#1{}
    
\fi

\ifcomments
    \usepackage{ulem}
    \def\mhedit#1{{$\!$\color{orange} [MH: #1]}}
    
\else 
    \def\mhedit#1{}
    
\fi

\begin{document}


\title{Non-Parametric and Regularized Dynamical Wasserstein Barycenters for \rev{Sequential Observations}}


\author{\IEEEauthorblockN{Kevin C. Cheng\IEEEauthorrefmark{1}, ~\IEEEmembership{IEEE Student Member}, Eric L. Miller\IEEEauthorrefmark{1}~\IEEEmembership{IEEE Fellow},\\ Michael C. Hughes\IEEEauthorrefmark{2}, Shuchin Aeron\IEEEauthorrefmark{1}~\IEEEmembership{IEEE Senior Member}}\\

\thanks{
\IEEEauthorrefmark{1} Tufts University, Dept. of Electrical and Computer Engineering\\
\indent\IEEEauthorrefmark{2} Tufts University, Dept. of Computer Science\\
This research was sponsored by the U.S. Army DEVCOM Soldier Center under the Measuring and Advancing Soldier Tactical Readiness and Effectiveness program and Cooperative Agreement Number W911QY-19-2-0003. We also acknowledge support from the U.S. National Science Foundation under award HDR-1934553 for the Tufts T-TRIPODS Institute. Shuchin Aeron is supported in part by NSF CCF:1553075, NSF RAISE 1931978, NSF ERC planning 1937057, and AFOSR FA9550-18-1-0465. Michael C. Hughes is supported in part by NSF IIS-1908617. Eric L. Miller is supported in part by NSF grants 1934553, 1935555, 1931978, and 1937057.
}
}

\IEEEtitleabstractindextext{%
\begin{abstract}
We consider probabilistic models for sequential observations which exhibit gradual transitions among a finite number of states. We are particularly motivated by applications such as human activity analysis where observed accelerometer time series contains segments representing distinct activities, which we call \textit{pure states}, as well as periods characterized by continuous transition among these pure states.  To capture this transitory behavior, the dynamical Wasserstein barycenter (DWB) model of \cite{cheng_dynamical_2021} associates with each pure state a data-generating distribution and models the continuous transitions among these states as a Wasserstein barycenter of these distributions with dynamically evolving weights.  Focusing on the univariate case where Wasserstein distances and barycenters can be computed in closed form, we extend \cite{cheng_dynamical_2021} specifically relaxing the parameterization of the pure states as Gaussian distributions.  We highlight issues related to  the uniqueness in identifying the model parameters as well as uncertainties induced when estimating a dynamically evolving distribution from a limited number of samples. To ameliorate non-uniqueness, we introduce regularization that imposes temporal smoothness on the dynamics of the barycentric weights.  A quantile-based approximation of the pure state distributions yields a finite dimensional estimation problem which we numerically solve using cyclic descent alternating between updates to the pure-state quantile functions and the barycentric weights.
We demonstrate the utility of the proposed algorithm in segmenting both simulated and real world human activity time series.
\end{abstract}

}

\maketitle

\IEEEdisplaynontitleabstractindextext

\IEEEpeerreviewmaketitle



\section{Introduction}

We consider a probabilistic model for  sequentially observed data where the observation at each point in time depends on a dynamically evolving latent state. We are particularly motivated by  systems that continuously move among a set of canonical behaviors, which we call \textit{pure states}. Over some periods, the system may reside entirely in one of the pure states while over other periods, the system is transitioning among these pure states in a temporally smooth manner. There are many applications where such a model is appropriate including climate modeling \cite{mukhin_principal_2015}, sleep analysis \cite{imtiaz_systematic_2021}, simulating physical systems \cite{chi_dynamical_2022}, as well as characterizing human activity from video \cite{hughes_nonparametric_2012} or wearable-derived accelerometry \cite{lara_survey_2012} data. Using the last case as an example, there will be periods when the individual will be engaged in a well-defined activity such as standing or running. During these intervals, the data can be modeled as drawn from a probability distribution specific to that canonical state. Given the high sampling rates of modern sensors, there also may be intervals where multiple consecutive observations reflect the gradual transition between or among pure states.  Over these periods the distribution of the data is given by a suitable combination of the pure state distributions. Therefore, one possible model for these types of systems consists of three components: a set of distributions containing the data-generating distribution for each pure state, a continuously evolving latent state which captures the transition dynamics of the system as it moves among these pure states, and a means of interpolating among these pure state distributions to characterize the data distribution in the transition regions. 

{\let\thefootnote\relax\footnotetext{Code repository: \url{https://github.com/kevin-c-cheng/DWB_Nonparametric}} }
{\let\thefootnote\relax\footnotetext{\copyright 2023 IEEE.  Personal use of this material is permitted.  Permission from IEEE must be obtained for all other uses, in any current or future media, including reprinting/republishing this material for advertising or promotional purposes, creating new collective works, for resale or redistribution to servers or lists, or reuse of any copyrighted component of this work in other works. DOI: 10.1109/TSP.2023.3303616}}

\begin{figure*}
 \centering
    \includegraphics[width=1.0\textwidth]{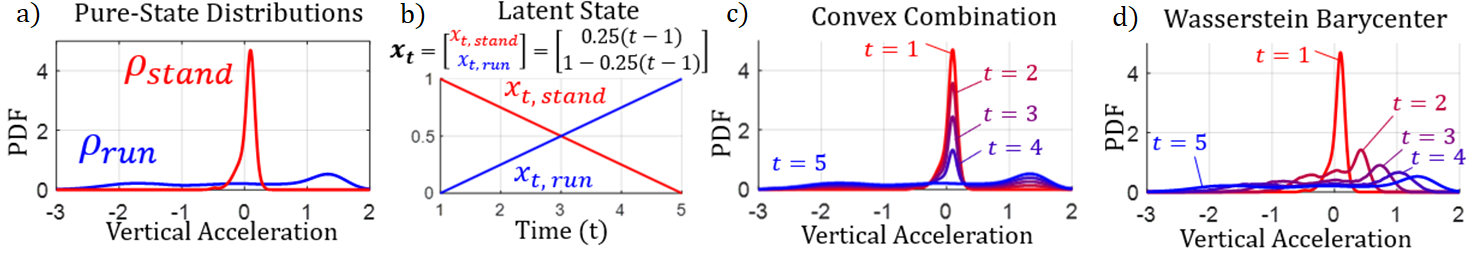}
  \caption{\textbf{Comparison of convex combination vs Wasserstein barycenter for modeling human activity transitions}. The beep test (BT) dataset consists of a subject running back-and-forth between two points, stopping at each point (see  Sec.~\ref{sec:RealWorldResults} for more details). (a) \rev{The probability distribution functions (PDF)} of the vertical acceleration of the system's two pure states (stand, run) are estimated using a KDE with a Gaussian kernel whose mean corresponds to the observed data when the system resides in these pure states. Modeling a transition from stand to run via the time-varying weights (b) for $t=1,..,5$, we show the resulting data distributions during this transition region according to a convex combination (c) and Wasserstein barycenter (d) interpolation model.}  \label{fig:MixtureVsWassBary}
\end{figure*}

These types of systems pose some unique considerations that are not sufficiently addressed by prior work in  time series modeling. The two most common methods for modeling latent state systems are continuous and discrete state-space models. Continuous state-space models \cite{kalman_new_1960, ribeiro_kalman_2004, fox_nonparametric_2008} have no natural way to identify those pure states in which the system may persist for periods of time. In discrete state-space models such as hidden Markov models, \cite{rabiner_tutorial_1989, ghahramani_factorial_1995, fox_joint_2014}, the dynamics are captured by a temporally varying state vector whose elements represent the probability that the system resides in each of a countable number of discrete (or in our terminology, pure) states.   For these models, the data-generating distribution associated with this latent state vector is given as a convex combination, i.e. a linear mixture, of the pure state distributions.  As argued in \cite{cheng_dynamical_2021} this is also an insufficient model for the problems which interest us.  As an example, for modeling human activity, the data distribution illustrated in Fig.~\ref{fig:MixtureVsWassBary} produced by a convex combination of the underlying ``standing'' and ``running'' pure state distributions can be interpreted as ``sometimes standing'' and ``sometimes running,'' which is not a proper description of the gradual transition that actually occurs. 

A better model for interpolating the data distribution during the transition period between standing and running would smoothly shift probability mass between the two pure state distributions. As illustrated in Fig.~\ref{fig:MixtureVsWassBary}, such blending can be achieved through displacement interpolation \cite{mccann_convexity_1997} between two pure states or, for more than two pure states, as a \emph{Wasserstein barycenter} of the associated distributions \cite{agueh_barycenters_2011}. Using this perspective, Dynamical Wasserstein Barycenters (DWB) \cite{cheng_dynamical_2021} were recently proposed to model a dynamically evolving distribution as a sequence of Wasserstein barycenters constructed as a time-varying convex combination of the pure state distributions. The dynamical weights, which lie on the probability simplex, are taken to be the latent state of the model.  A Bayesian model is proposed in \cite{cheng_dynamical_2021}, whose parameters were determined via maximum \textit{a posteriori} estimation.

Here we expand on \cite{cheng_dynamical_2021} by highlighting two challenging characteristics of the DWB model and two improvements that address certain limitations of the original DWB approach. The first characteristic relates to uniqueness in DWB model identification, where multiple combinations of pure state distributions and barycentric weights can produce the same Wasserstein barycenter.  
Although this is true for multidimensional distributions, here we use a univariate formulation to more transparently demonstrate how this non-uniqueness is captured in an \textit{inverse-scaling} relationship between the model's latent state and pure state parameters. The second characteristic is related to a tradeoff in tracking and estimating an evolving data distribution from a single instance of a time series using a windowed approach to collect samples.  Smaller windows lack the number of samples to ensure small statistical error in the estimation of the data distribution at a given point in time. 
On the other hand, larger windows span longer periods during which, under relatively faster dynamics, the data distribution can change significantly again increasing the estimation error. In a simulated example, where the dynamics consist of constant rate transitions between two Gaussian states, we show that there exists an optimal window size that balances these two effects and discuss the dependency of this window size tradeoff on the temporal dynamics of the latent state and pure states of the system.

Our first improvement addresses the limitations of the choice in \cite{cheng_dynamical_2021} in using a probabilistic prior for the dynamics of the DWB weight vector. That approach may introduce additional unnecessary or potentially undesirable probabilistic properties on the latent state process such as a limiting distribution  \cite{nguyen_class_2020} which fails to adequately regularize the DWB estimation problem. Instead, we propose here a regularization scheme that imposes temporal smoothness by penalizing the difference between the simplex-constrained, latent state vectors at adjacent points in time. Drawing from the field of compositional analysis \cite{martin-fernandez_measures_1998}, the Bhattacharya-arccos distance \cite{astrom_image_2017} proves to be well-suited to our needs.  As a consequence of the aforementioned inverse-scaling relationship, introducing this latent state regularizer impacts the model's pure state distribution in a manner that causes them to diverge from the data. Therefore, we also introduce a regularizer to counteract this effect to ensure that the learned pure state distributions are representative of data while the system resides in each pure state.

Our second improvement removes the restriction in \cite{cheng_dynamical_2021} where a parametric approach to model pure states with multivariate Gaussians was employed. Here we adopt a non-parametric approach and focus on the univariate case where the Wassertein-2 distance between distributions is equivalent to the 2-norm between their respective quantile functions \cite{peyre_computational_2019}. Using a discrete approximation to the pure state quantile functions leads to a convenient finite dimensional, regularized linear least squares problem for estimating the pure states.

Our numerical experiments empirically validate our analysis and improvements to the DWB model.  Using simulated data, we demonstrate in a controlled setting how we effectively regularize our model parameters with proper consideration of the inverse-scaling analysis and the impact of window size on the accuracy of the model parameters. Additionally, using real world human activity data, we show how our non-parametric approach leads to improved estimation of the system's pure state distributions as well as improved fit of the time-evolving distribution of the observed data compared to  \cite{cheng_dynamical_2021}.

In summary, the primary contributions of this work consist of the following:
\begin{enumerate}
    \item We highlight the non-uniqueness of the parameters corresponding to a Wasserstein barycenter by detailing the inverse-scaling relationship between the pure state distributions and the simplex-valued barycentric weights.
    
    \item We explore the impact of the window size on the ability to accurately estimate a dynamically evolving data distribution by exploring the tradeoff between the errors associated with large and small windows and the dependency of this tradeoff on the dynamics and pure states of the system.
        
    \item We propose regularizers for the model parameters that impose temporal smoothness in the latent states in a manner that addresses the non-uniqueness of the model. 

    \item We propose a flexible, non-parametric representation for univariate pure state distributions using a discrete approximation to the quantile function that results in a finite dimensional formulation for DWB learning.
\end{enumerate}

The remainder of the paper is organized as follows: in Sec.~\ref{sec:TechnicalBackground}, we provide an overview of the Wasserstein distance and barycenter focusing on the univariate case. In Sec.~\ref{sec:DWB_Model}, we discuss the DWB model, highlighting the non-uniqueness and inverse-scaling property of the Wasserstein barycenter as well as the impact of the window size on the estimation of a dynamically evolving data distribution. In Sec.~\ref{sec:NewDWB_Model}, we develop a variational problem for learning a DWB model, followed by a discussion of the regularization approach, and discretization of the pure state distributions required to obtain a finite dimensional estimation problem. We then formally state our non-parametric and regularized DWB variational problem and provide an algorithm to estimate the model parameters. In Sec.~\ref{sec:Evaluation} we use simulated data to demonstrate the non-uniqueness, impact of window size and regularization terms discussed in this work and use real world human activity data to demonstrate the advantages of the non-parametric DWB approach relative to the Gaussian model.

\section{Technical Background}
\label{sec:TechnicalBackground}
The Wasserstein-2 distance is a metric on the space of probability distributions on $\mathbb{R}^d$ with finite second moments \cite{peyre_computational_2019, santambrogio_optimal_2015}. For two random variables $q$ and $s$ distributions $\meas_q$ and $\meas_s$, the squared Wasserstein-2 distance is defined via,
\begin{align}\label{eq:2Wass}
        \mathcal{W}_2^2(\meas_q,\meas_s) = \inf_{\pi\in \Pi(\meas_q,\meas_s)} \mathbb{E}_{q, s \sim \pi}\|q-s\|_2^2
\end{align}
where $\pi$ denotes the joint distribution of $q$ and $s$, and $\Pi(\meas_q, \meas_s)$ is the set of all joint distributions with marginals $\meas_q, \meas_s$. In this work, we refer to Eq.~\eqref{eq:2Wass} as the squared Wasserstein distance. 

Given a set of distributions $\meas_{q_{1:K}} = \{\meas_{q_1}, \meas_{q_2}, ..., \meas_{q_K}\}$ and a vector $\xmb \in \Delta^K$, where $\Delta^K$ denotes the standard $K$-simplex, the \textit{Wasserstein barycenter} is the distribution that minimizes the weighted (with respect to elements in $\xmb$) squared Wasserstein distance to the set of distributions \cite{agueh_barycenters_2011} and is given by,
\begin{align} \label{eq:Barycenter}
    \meas_{B} =& B(\xmb, \meas_{q_{1:K}}) = \argmin_\meas \sum_{k=1}^{K} \xmb[k] \mathcal{W}_2^2 (\meas, \meas_{q_k}),
\end{align}
\rev{where $\xmb[k]$ denotes the $k$-th element of the vector $\xmb$.} 
When $\meas_q$ and $\meas_s$ are univariate distributions with cumulative distribution functions $P_q, P_s$, the squared Wasserstein distance in Eq.~\eqref{eq:2Wass} becomes \cite{santambrogio_optimal_2015, bonneel_sliced_2013}, 
\begin{align} \label{eq:1dWass}
    \mathcal{W}_2^2(\meas_q,\meas_s) = \int_0^1 \left( P_q^{-1}(\xi)-P_s^{-1}(\xi) \right)^2 d\xi.
\end{align}
Here $P_q^{-1}$ and $P_s^{-1}$ are quantile functions, the generalized inverse \cite{embrechts_note_2013} of the cumulative distribution function, given by,
\begin{align}
    \QuantP(\xi) = \inf \{ g \in \mathbb{R}: P(g) \geq \xi\}.
\end{align}
It follows from Eq.~\eqref{eq:1dWass} and Eq.~\eqref{eq:Barycenter} that the Wasserstein barycenter of a set of univariate distributions with quantile functions $P_{q_{1:K}}^{-1}$, will have quantile function \cite{bonneel_sliced_2013},
\begin{align} \label{eq:1dWassBary}
    P_B^{-1} =& \sum_{k=1}^K \xmb[k] P_{q_k}^{-1}.
\end{align}

\section{The Dynamical Wasserstein Barycenter Model} \label{sec:DWB_Model}
\begin{figure}
 \centering
    \includegraphics[width=0.489\textwidth]{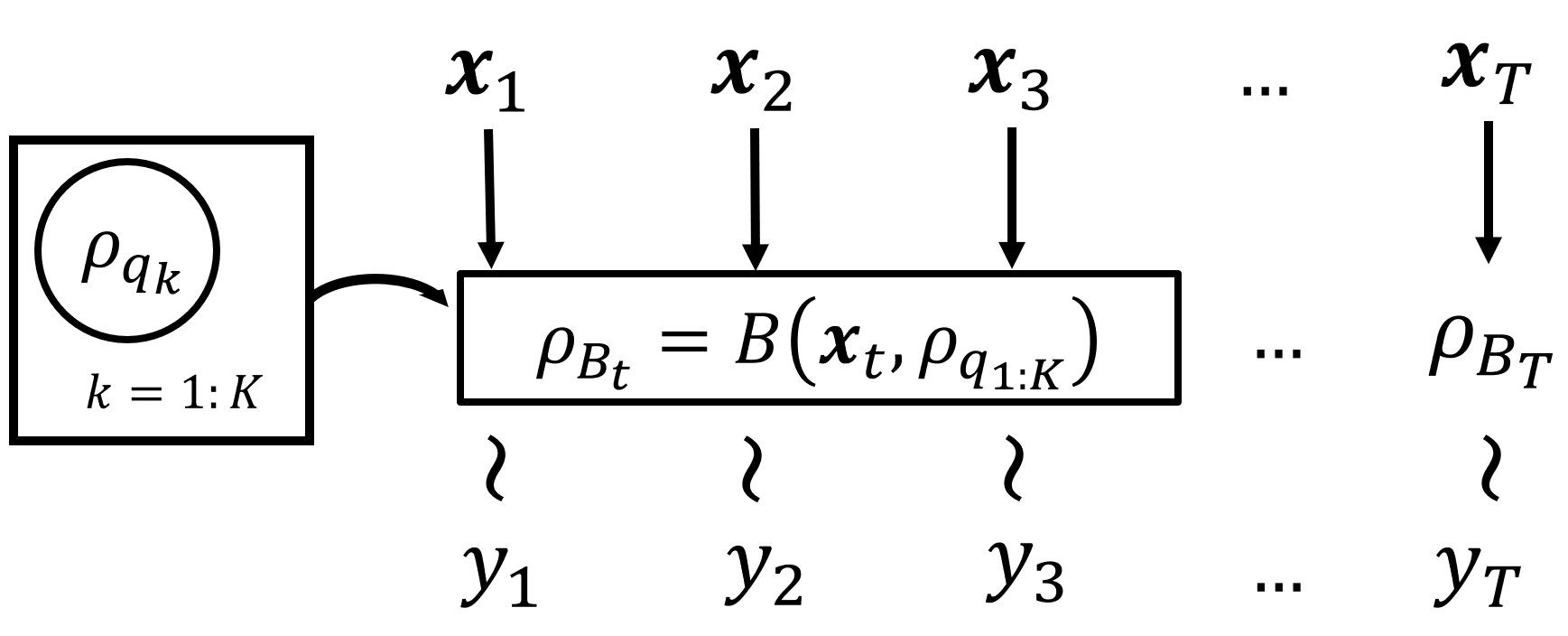}
  \caption{\textbf{DWB model diagram}. The DWB models the distribution $\meas_{B_t}$ from which the time series $y_t$ is sampled as the Wasserstein barycenter of a set of pure state distributions $\meas_{q_{1:K}}$ and barycentric weight $\xmb_{1:T}$, the latent state of the model. }  \label{fig:DWB_Diagram}
\end{figure}
Shown in Fig.~\ref{fig:DWB_Diagram}, the DWB model \cite{cheng_dynamical_2021} describes the  distribution of a time series $y_t$ at time $t$ as,
\begin{align} \label{eq:ModelBarycenter}
    y_t \sim \meas_{B_t} = B(\xmb_t,\meas_{q_{1:K}})
\end{align}
where $\meas_{q_k}$, $k = 1,2,\dots,K$ are the distributions of the pure states and the barycentric weight $\xmb_t \in \Delta^K$ capture the dynamics of the transitions among these pure states. 

Given $y_t, t = 1,2,\cdots, T$ modeled via equation~\eqref{eq:ModelBarycenter}, the problem is to estimate DWB model parameters which consist of the pure state distributions and the sequence of barycentric weights. 

Below we discuss two key characteristics that pose challenges for estimating the parameters of the DWB model. The first is the non-uniqueness of the parameters (i.e., the pure state distributions and the barycentric weights) that yield a Wasserstein barycenter.  The second relates to the complications that arise when we are provided only a single time series for learning a DWB model.

\subsection{Non-uniqueness in the Parameters of a Wasserstein Barycenter}  \label{sec:Uniqueness}
\begin{figure*}
 \centering
    \includegraphics[width=1.0\textwidth]{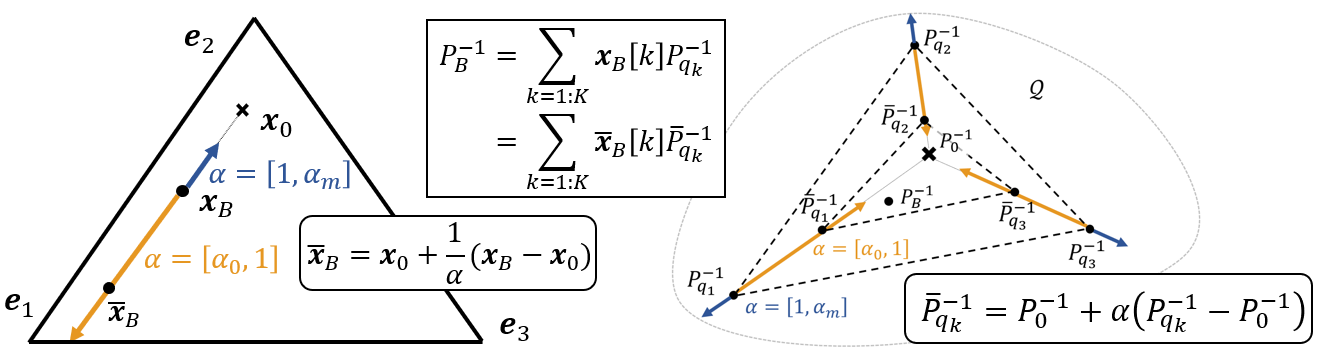}
  \caption{\textbf{Diagram of the non-uniqueness and inverse-scaling effect of the parameters of a Wasserstein barycenter.} Consider a set of three pure states with quantile functions $\QuantP_{q_{1:3}}$ and simplex-valued weight $\xmb_B\in \Delta^3$ where $\meas_B=B(\xmb_B, \meas_{q_{1:3}})$ with quanitle function $\QuantP_B = \sum_{k=1}^3 \xmb_B[k]\QuantP_{q_k}$. We construct a family of distinctly different pure state quantile functions $\bar{P}_{q_{1:K}}^{-1}$ and barycentric weights $\bar{\xmb}_B$ which produce the exact same barycenter $\QuantP_B = \sum_{k=1}^3 \bar{\xmb}_B[k]\bar{P}_{q_k}^{-1}$. Let $\xmb_0$ be another point on the simplex where $\meas_0 = B(\xmb_0, \meas_{q_{1:3}})$ has quantile function $\QuantP_0 = \sum_{k=1}^3 \xmb_0[k]\QuantP_{q_k}$. 
  Given $\xmb_0$ and $\xmb_B$, let $\bar{\xmb}_B$ given by Eq.~\eqref{eq:UniquenessX} be any point on the line connecting $\xmb_0$ through $\xmb_B$ to the edge of the simplex. Moving $\bar{\xmb}_B$ \textit{away} from $\xmb_0$ along the line connecting $\xmb_0$ and $\xmb_B$ (orange segments), causes the pure states quantile functions $\bar{P}^{-1}_{q_{1:3}}$ to move from $\QuantP_{q_{1:3}}$ \textit{towards} $\QuantP_0$. This corresponds to $\alpha \in [\alpha_0, 1]$ where $\alpha_0$ is the smallest value of $\alpha$ such that $\bar{\xmb}_B$ still lies on the simplex. Conversely moving $\xmb_B$ \textit{towards} $\xmb_0$ (blue segments) results in the pure state quantile functions moving \textit{away} from $\QuantP_0$. This corresponds to $\alpha \in [1, \alpha_m]$, where $\alpha_m$ is the largest value of $\alpha$ such that all $\bar{P}_{q_{1:3}}$ remain in the set of quantile functions $\mathcal{Q}$.} \label{fig:Uniqueness}
\end{figure*}

The issue of uniqueness refers to the fact that a Wasserstein barycenter is not described by a unique set of pure state distributions and barycentric weights.  While the statement is true regardless of dimension (see Appendix~\ref{sec:Appendix:Uniqueness} for an example), given the focus of this paper, we examine the univariate case in some detail. Specifically, we provide a construction that illustrates an inverse-scaling relation between the family of pure state distributions and barycentric weights that yields the same Wasserstein barycenter.

As shown in Fig.~\ref{fig:Uniqueness}, assume we have a set of pure state distributions $\meas_{q_{1:K}}$ indexed by $k=1,2,\dots, K$ with quantile functions, $P_{q_k}^{-1}$ and barycentric weights $\xmb_B\in \Delta^K$ which give rise to the barycenter $\meas_B=B(\xmb_B, \meas_{q_{1:K}})$ with quantile function $P_B^{-1} = \sum_{k=1}^K\xmb_B[k]P_{q_k}^{-1}$ (Eq.~\eqref{eq:1dWassBary}). For now, $\xmb_B$ is assumed to lie in the interior of the simplex and we consider below the cases where $\xmb_B$ is on a lower dimensional face or vertex.  Let us choose another point $\xmb_0 \neq \xmb_B$ corresponding to barycentric quantile function $P_0^{-1} = \sum_{k=1}^{K}\xmb_0[k]P_{q_k}^{-1}$ (Eq.~\eqref{eq:1dWassBary}). We construct a family of barycentric weights $\bar{\xmb}_B$ and pure state quantile functions, $\bar{P}^{-1}_{q_{1:K}}$ corresponding to distributions $\bar{\meas}_{q_{1:K}}$ such that $\meas_B=B(\bar{\xmb}_B, \bar{\meas}_{q_{1:K}})$, or in other words, $P_B^{-1} = \sum_{k=1}^K\bar{\xmb}_B[k]\bar{P}_{q_k}^{-1}$.
Specifically, as seen in Fig.~\ref{fig:Uniqueness} we define $\bar{\xmb}_B$ to be a point on the line segment connecting $\xmb_0$ to the boundary of the simplex that passes through $\xmb_B$. That is, 
\begin{align} \label{eq:UniquenessX}
\bar{\xmb}_B = \xmb_0 + \frac{1}{\alpha}(\xmb_B -\xmb_0).
\end{align}
The parameter $\alpha$ captures the scaling nature of this construction. When $\alpha = \infty$ we have $\bar{\xmb}_B = \xmb_0$. As $\alpha$ decreases, $\bar{\xmb}_B$ moves away from $\xmb_0$ along the blue segment connecting $\xmb_0$ and $\xmb_B$, ultimately crossing $\xmb_B$ when $\alpha = 1$. Further reducing $\alpha$ towards $0$ moves $\bar{\xmb}_B$ along the orange component of the same line until some point $\alpha \in (0,1]$, where $\bar{\xmb}_B$ reaches the boundary. We denote this point as $\alpha_0$.

With this definition of $\bar{\xmb}_B$ in Eq.~\eqref{eq:UniquenessX} we construct quantile functions, $\bar{P}_{q_{1:K}}^{-1}$ such that $P_B^{-1} = \sum_{k=1}^K\bar{\xmb}_B[k]\bar{P}_{q_k}^{-1}$. Indeed,
\begin{align}
    &\QuantP_B = \sum_{k=1}^{K} \xmb_B[k] \QuantP_{q_k} 
    = \sum_{k=1}^{K} \left( \alpha \bar{\xmb}_B[k]  + (1-\alpha) \xmb_0[k] \right)\QuantP_{q_k}\nonumber \\
    &= \alpha \sum_{k=1}^{K} \bar{\xmb}_B[k] \QuantP_{q_k}+
       (1-\alpha)\sum_{k=1}^{K} \xmb_0[k]\QuantP_{q_k} \nonumber \\  
    &= \alpha \sum_{k=1}^{K} \bar{\xmb}_B[k] \QuantP_{q_k} + (1-\alpha)\QuantP_0  
    \hspace{5mm}
    \boxed{P_0^{-1} = \sum_{k=1}^K\xmb[k]P_{q_k}^{-1}} \nonumber\\
    &= \sum_{k=1}^{K}\left( \alpha  \bar{\xmb}_B[k] \QuantP_{q_k} + (1-\alpha) \bar{\xmb}_B[k]\QuantP_0 \right)  
    \hspace{2mm}
    \boxed{\sum_{k=1}^K\bar{\xmb}_B[k] = 1} \nonumber\\
    &= \sum_{k=1}^{K} \bar{\xmb}_B[k]\left( \alpha \QuantP_{q_k} + (1-\alpha)\QuantP_0 \right) \nonumber
    = \sum_{k=1}^{K} \bar{\xmb}_B[k]\bar{P}^{-1}_{q_k}
\end{align}
with for each $k$,
\begin{align} \label{eq:UniquenessP}
    \bar{P}^{-1}_{q_k} = \QuantP_0 + \alpha (\QuantP_{q_k}-\QuantP_0).
\end{align}
Eq.~\eqref{eq:UniquenessP} bears a strong resemblance to Eq.~\eqref{eq:UniquenessX}, except now with reciprocal use of $\alpha$. For $\alpha \in [\alpha_0,1]$, $\bar{P}^{-1}_{q_k}$ is a convex combination of $\QuantP_0$ and $\QuantP_{q_k}$ lying on the line segment connecting the two quantile functions. In this case, since the collection of monotone functions on $[0,1]$ is a convex set \cite{rakestraw_convex_1972},  $\bar{P}^{-1}_{q_{1:K}}$ will be valid quantile functions. However, for  $\alpha>1$, $\bar{P}^{-1}_{q_k}$ extends beyond $\QuantP_B$ along the line that connects $\QuantP_0$ to $\QuantP_B$. In this case, $\bar{P}^{-1}_{q_k}$ is no longer a convex combination of $\QuantP_0$ and  $\QuantP_B$ and is not guaranteed to be a quantile function. We denote $\alpha_m$ as the maximum value of $\alpha$ such that $\bar{P}^{-1}_{q_k}$  is a valid quantile function.

Thus, the sets of $\bar{\xmb}_B$ and  $\bar{P}^{-1}_{q_{1:K}}$ corresponding to $\alpha \in [\alpha_0, \alpha_m]$ according to Eqs.~\eqref{eq:UniquenessX} and \eqref{eq:UniquenessP} describe the family of parameters that yield the same Wasserstein barycenter as $\xmb_B$ and $P^{-1}_{q_{1:K}}$. The reciprocal appearance of $\alpha$ in these equations captures the inverse-scaling relationship for this family of parameters. As shown in Fig.~\ref{fig:Uniqueness}, the case where $\alpha \in [\alpha_0,1]$ corresponds to the orange lines where \textit{increasing} the distance of $\bar{\xmb}_B$ from $\xmb_0$ along the segment connecting $\xmb_B$ and $\xmb_0$, results in the pure states \textit{decreasing} their distances to $P_0^{-1}$ each along linear trajectories connecting the $P_{q_k}^{-1}$ to $P_0^{-1}$.  
Conversely, the case where $\alpha \in [1, \alpha_m]$ corresponds to the blue lines in Fig.~\ref{fig:Uniqueness} where \textit{decreasing} the distance between $\bar{\xmb}_B$ and $\xmb_0$ by moving along the line that connects $\xmb_B$ to $\xmb_0$, results in pure state quantile functions $\bar{P}_{q_k}^{-1}$ that are now \textit{increasing} their distance from $P_0^{-1}$ by extending linearly along the ray from $\QuantP_0$ to $\QuantP_{q_k}$. 

Should $\xmb_B$ lie on a face of the $K$-simplex of dimension greater than one but less than $K$ (i.e.,~\textit{not} a vertex), this construction may be repeated by placing $\xmb_0 \neq \xmb_B$ in that same lower dimensional simplex. In such a case, we may also place $\xmb_0$ in the interior of the $K$ dimensional simplex.  More specifically, with $\xmb_B$ on a face and $\xmb_0$ located in the interior, it is clear that $\alpha_0 = 1$ implying that the construction above holds only if $\alpha_m > 1$. This is also the case in the event that $\xmb_B$ is a vertex for any value of $\xmb_0 \neq \xmb_B$. With the constraint that $\bar{P}^{-1}_{q_{1:K}}$ must remain valid quantile functions, it is possible to construct an example such that $\alpha_m=1$, which when combined with the aforementioned case where $\alpha_0=1$ means that we cannot employ this construction to show non-uniqueness. However, as discussed in Appendix \ref{sec:Appendix:Uniqueness} neither can we conclude that the barycenter is in fact unique.

This non-uniqueness and inverse-scaling relation implies that for models (such as the DWB) that require learning both the pure state quantile functions and the barycentric weights corresponding to one or a sequence of Wasserstein barycenters, introducing constraints on one set of parameters will have an impact on the other.  We take this effect into account in Sec.~\ref{sec:Regularization} when constructing regularizers to impose desirable properties on the latent state sequence and pure state distributions.

\subsection{Model Sampling} \label{sec:Windowing}
A second characteristic of learning a DWB model relates to how we incorporate observed data into the model. Ideally, we would directly observe the data distribution $\meas_{B_t}$ as specified by Eq.~\eqref{eq:ModelBarycenter} at each point in time or generate an estimate of these distributions from multiple realizations of a time series. Unfortunately, in all the practical cases of interest to us, only a single instance of the time series is available for processing.  Thus, we consider the problem of estimating the time-varying data distribution $\meas_{B_t}$ from a single time series that is sampled from $\meas_{B_t}$. To do this, we consider a window of $n$ samples centered at $t$, compiled into a vector $\ymb_t = [y_{(t-\frac{n}{2})},\dots, y_{(t+\frac{n}{2}-1})]^\TP$.  For convenience of notation, we assume $n$ to be even.  We estimate the data distribution with the distribution $\meas_{y_t}$ based on this sample window,
\begin{align}\label{eq:empirical}
    \meas_{y_t} = \frac{1}{n}\sum_{i=1}^{n} \delta_{\ymb_t[i]}.
\end{align}
Here, $\delta_{\ymb_t[i]}$ is the Dirac-delta measure located at $\ymb_t[i]$.
We then define the \textit{window approximation error} (WAE) as,
\begin{align}\label{eq:WindowApproximationError}
    e_{w_t} =\mathcal{W}_2^2 (\meas_{y_t}, \meas_{B_t}).
\end{align}
Since the samples that constitute $\meas_{y_t}$ are random, the WAE is a random quantity.
\begin{figure*}
 \centering
    \includegraphics[width=1.0\textwidth]{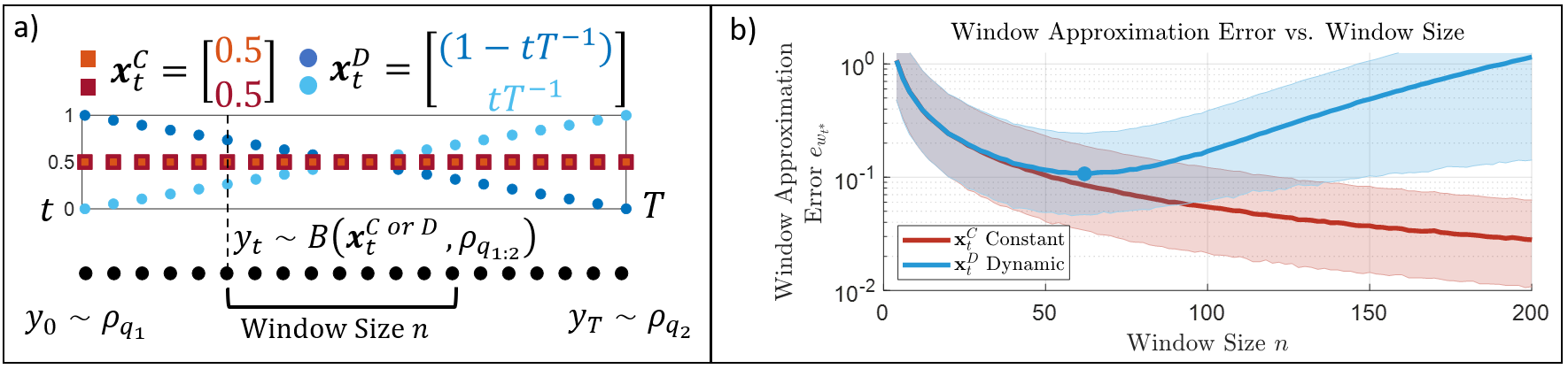}
  \caption{\textbf{Approximating a distribution with a window of samples}. (a) Two simulated configurations, one where $\xmb^C_t =[0.5,0.5]^\TP$ is constant, and one where $\xmb^D_t=[(1-tT^{-1}), tT^{-1}]^\TP$ is dynamically evolving. We generate a time series $y_t \sim B(\xmb_t^{C~\textnormal{or}~D}, \meas_{q_{1:2}})$ where $\meas_{q_1} = \mathcal{N}(0,5)$, and  $\meas_{q_2} = \mathcal{N}(10,0.2)$\protect\footnotemark.
  (b) We estimate the distribution data distribution $\meas_{B_{t^*}}$ where $t^*=0.5(T+1)$, with a distribution $\meas_{y_t^*}$ comprised of a window of $n$ samples centered at $t^*$ according to Eq.~\eqref{eq:empirical}. Varying the window size $n$ we plot the average value and the 25/75-th quantile bands of $e_{w_{t^*}} = \mathcal{W}_2^2(\meas_{y_{t^*}}, \meas_{B_{t^*}})$ from $10^4$ simulations. In the constant state case where $\meas_{y_t^*}$ consists of $n$ IID samples from $\meas_{B_t^*}$ the average error monotonically decreases. However, in the dynamic case, where the samples in the window are independent but not identically distributed, the U-shape curve highlights the window size tradeoff where $n_0$ indicates the optimal window size. }  \label{fig:CVD}
\end{figure*}
\footnotetext{$\mathcal{N}(\mu, \sigma^2)$: Gaussian distribution with mean $\mu$, and variance $\sigma^2$.}

\begin{figure}
 \centering
    \includegraphics[width=0.489\textwidth]{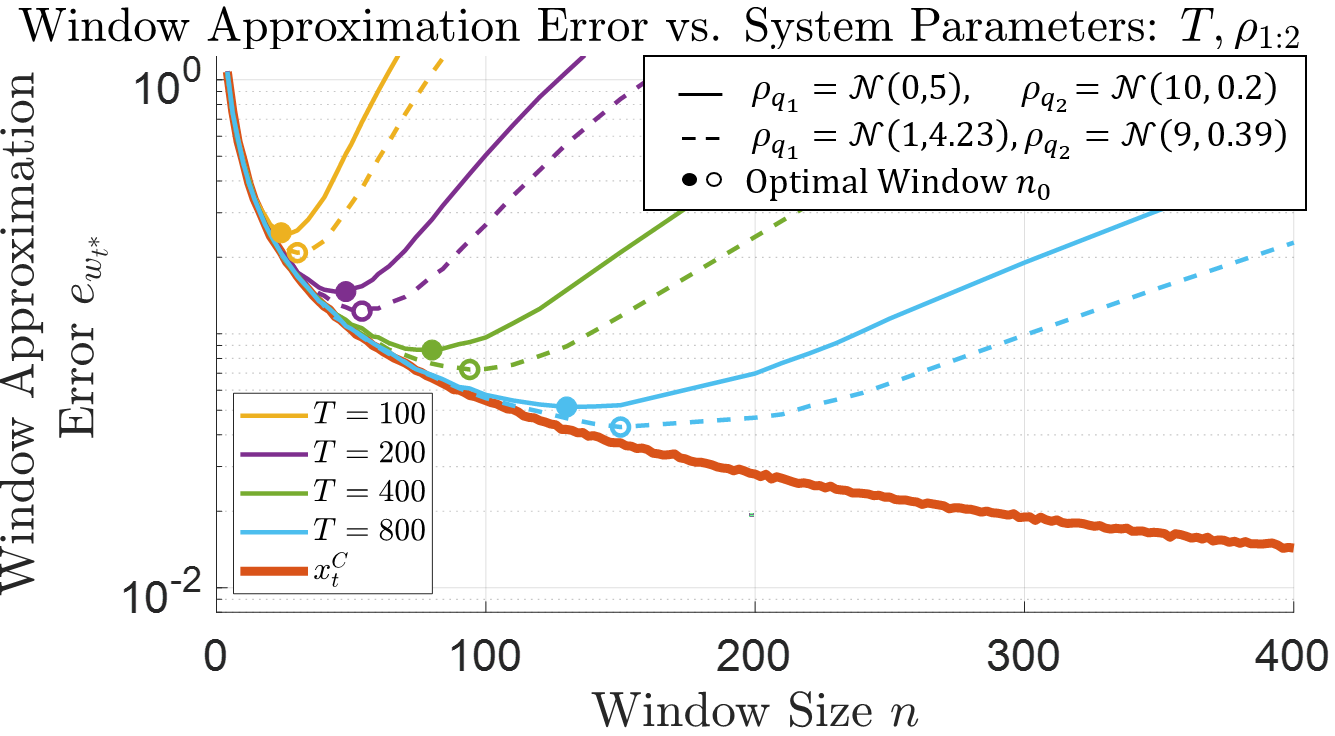}
  \caption{\textbf{Impact of system dynamics and pure state distributions on WAE}.
  The average WAE is plotted for various configurations where solid lines refer to $\meas_{q_1}=\mathcal{N}(0,5)$,  $\meas_{q_2}=\mathcal{N}(10,0.2)$ and dashed lines corresponds to a system where the Wasserstein distance between $\meas_{q_{1:2}}$ is decreased, $\meas_{q_1}=\mathcal{N}(1,4.23)$ $\meas_{q_2}=\mathcal{N}(9,0.39)$. In both cases, $\meas_{B_{t^*}}=\mathcal{N}(5,1.8)$.
  Decreasing the per-sample change in distribution by increasing $T$ or decreasing $\mathcal{W}_2^2(\meas_{q_1},\meas_{q_2})$ (solid $\rightarrow$ dashed) results in smaller $e_{w_{t^*}}$. 
  The impact of these changes on $e_{w_{t^*}}$ is greater for larger windows, which increases the optimal window $n_0$ corresponding to the minimum of the U-curves.}  \label{fig:WindowExperimentResults}
\end{figure}

Eq.~\eqref{eq:empirical} is an empirical measure when the samples in the window are drawn IID which is only possible when the barycentric weights $\xmb_t$, and thus $\meas_{B_t}$, is constant in time.  In this case, as $n \rightarrow \infty$, it is known that the distribution function of $\meas_{y_t}$ converges almost surely to the distribution function of $\meas_{B_t}$, and the expected value of $e_{w_t}$ converges to zero \cite{fournier_rate_2015}. The complication in our case comes from the fact that $\xmb_t$ is changing with time, meaning that the samples in the window are no longer identically distributed as the data distribution changes from sample to sample.

This impact of a dynamically evolving distribution on the WAE is dependent on many factors including the size of the window $n$ as well as properties of the system that impact the manner in which the data distribution changes between samples. To simplify matters, we consider a simple yet informative example in Fig.~\ref{fig:CVD}a of a system with two pure states and two latent state configurations, one where the latent state is \textit{constant} with $\xmb^C_t=[0.5, 0.5]^\TP$, and one where the latent state is \textit{dynamic}, changing at a constant rate where $\xmb^D_t = [(1-tT^{-1}), tT^{-1}]^\TP$, for $t=0,1,\dots,T$.
For two distributions $\meas_{q_{1:2}}$, we generate a time series by sampling independently $y^{C \text{ or }D}_t \sim B(\xmb^{C \text{ or }D}_t, \meas_{q_{1:2}})$. Thus, in the constant state case, $y^C_t$ are IID.  Let $\meas_{y_{t^*}}$ be the empirically estimated  distribution constructed according to Eq.~\eqref{eq:empirical} at $t^*=0.5(T+1)$, the half-way point of the transition. In both cases of $\xmb^C_t$ and $\xmb^D_t$ the distribution at this time is $\meas_{B_{t^*}} = B([0.5, 0.5]^\TP, \meas_{q_{1:2}})$.

When the window size is small, the distributional change over the sample window is also small. Therefore as seen in the left side of Fig.~\ref{fig:CVD}b, the dynamic case closely approximates the constant state where the average WAE decreases with respect to the window size \cite{fournier_rate_2015}. 
When the window size is large, the samples being included in the window are further from $t^*$ and thus the distributions of these samples increasingly diverge from $\meas_{B_{t^*}}$. Shown by the rising and expanded quantile bands on the right side of the plot, this effect causes the average and variability of $e_{w_{t^*}}$ to increase for large windows. The resulting U-shape curve of the WAE highlights the window size tradeoff where the optimal window size $n_0$ corresponding to the minimum of this curve balances the benefits of having more samples for robust estimation of a distribution with the effects of using samples farther from the point of interest.

One major factor that impacts this tradeoff is the magnitude of the distributional change from sample to sample. There are a number of factors that can decrease (resp. increase) the magnitude of this per-sample change in distribution including (1) decreasing (increasing) the rate of change of the system's continuous time dynamics; (2) increasing (decreasing) the sampling rate of the sensor which provides discrete measurements; or (3) decreasing (increasing) the Wasserstein distance between the pure state distributions which affects the total amount of distributional change during the transition among pure states. Continuing with the example, we demonstrate the impact of these factors on the window size tradeoff and optimal window size. Again, modeling the system as moving from $\meas_{q_1}$ to $\meas_{q_2}$ with dynamics specified by $\xmb_t^D$, factors (1) and (2) dictate the number of samples over which this transition occurs\footnote{The rate of change (1) may have units change in distribution per second, and the sampling rate (2) has units samples per second. Thus $\frac{(1)}{(2)}$ will have units change in distribution per sample.}, thus their combined impact can be understood by varying $T$. To understand the impact of (3), we simulate two different pure state configurations varying $\mathcal{W}_2^2(\meas_{q_1}, \meas_{q_2})$. 

The results in Fig.~\ref{fig:WindowExperimentResults} confirm that for a given window size, decreasing the per-sample change in distribution by increasing $T$, or by decreasing $\mathcal{W}_2^2(\meas_{q_1}, \meas_{q_2})$ results in a decrease in the average WAE. Additionally, the increasing difference between the U-curves in Fig.~\ref{fig:WindowExperimentResults} as we move towards larger windows confirms that the decreasing the per-sample change in distribution has an increasing benefit for larger windows where the dynamics of the system have a larger impact on the accuracy of the window estimate.
This shifts the balance of the window size tradeoff as seen by the minimums of these U-curves moving to the right, implying that decreasing the per-sample change in distribution using any of the three methods mentioned increases the optimal window size. Indeed, in the limiting case where either $T\rightarrow \infty$ or $\mathcal{W}_2^2(\meas_{q_1}, \meas_{q_2})\rightarrow 0$ in which case $\meas_{q_1}=\meas_{q_2}=\meas_{B_{t^*}}$, the samples will be drawn IID from a constant distribution. In this limit, the dynamic case converges to the constant case where the optimal window size $n_0 \rightarrow \infty$ and $e_{w_t} \rightarrow 0$.

\section{Non-Parametric and Regularized Dynamical Wasserstein Barycenters} \label{sec:NewDWB_Model}
In this section, we detail our proposed variational problem for estimating the parameters of the univariate DWB model. We discuss our regularization that ensures that the latent state evolves smoothly over time while taking into account its effect on the pure states through the inverse-scaling relationship. We also discuss our non-parametric representation for the pure state distributions using a discrete approximation to the pure state quantile function. Finally, we detail how this leads to a least-squares formulation of the variational DWB objective and propose an algorithm for learning the parameters of the model. 

\subsection{Variational Problem for DWB Model Estimation}\label{sec:Estimation}

Training a DWB model entails estimating the pure states distributions and latent states sequence to minimize a cost function that encourages both fidelity to the data as well as model parameters that conform to prior knowledge we may have concerning the general behavior of time series.

Building on the approach discussed in Sec.~\ref{sec:Windowing}, we create a sequence  of $\T$ sample windows of length $n$ to estimate the distribution of the time series at select points.  For our simulations in Sec.~\ref{sec:Evaluation} we use overlapping windows separated by a fixed stride length. Given $t_i$ for $i=1,\dots,\T$ as the starting index for these sample windows, let $\ymb_i =[y_{t_i}, ..., y_{{t_i}+n}]^\TP$ be the vector of samples and $\meas_{y_i}$ the distribution according to Eq.~\eqref{eq:empirical} corresponding to this window of samples. Using the WAE in Eq.~\eqref{eq:WindowApproximationError} summed over $i$ as a data fidelity term and encoding prior information in regularizers the details of which are discussed below, the variational problem we seek to solve is,
\begin{align} \label{eq:Loss}
    &\hat{\meas}_{q_{1:K}}, \hat{\xmb}_{1:\T} = \argmin_{\meas_{q_{1:K}}, \xmb_{1:\T}}  \nonumber \\
    & \hspace{2mm} \underbrace{\sum_{i=1}^{\T} \mathcal{W}_2^2 \left( \meas_{y_i}, \meas_{B_i} \right)}_{\text{Data Fit}}
            + \lambda_{x} \underbrace{ R_x\left( \xmb_{1:\T} \right)}_{\text{x-regularization}}
            + \lambda_{q}\T\underbrace{ R_{q}(\rho_{q_{1:K}})}_{\text{q-regularization}},
\end{align}
where $\meas_{B_i} = B(\xmb_i, \meas_{q_{1:K}})$. Here, $\lambda_{x}, \lambda_{q} \geq 0 $ are regularization weights. We multiply $R_{q}$ by $\T$ so that it scales with the length of the time series along with the other terms in the cost function.

As highlighted by the inverse-scaling relation in Sec.~\ref{sec:Uniqueness}, any regularizer of either the latent state or pure state must consider its effect on the other parameter. Motivated by applications where the system evolves gradually over time, we propose a regularization scheme that imposes a gradually evolving latent state while ensuring that the learned pure state distributions accurately reflect the distribution of the data corresponding to when the system resides in a pure state.

\subsection{Parameter Regularization for DWB} \label{sec:Regularization}
As seen in our windowing simulations in Sec.~\ref{sec:Windowing}, even in a simple case, estimating a dynamically evolving distribution with a window of samples is a challenging problem. The ambiguities identified in our experiments in Sec.~\ref{sec:Windowing} may cause the learned latent state to vary greatly even if the system is constant or gradually evolving. Therefore, to limit its variability, we propose a regularizer that penalizes the sum of the squared distances, $d^2(\xmb_i, \xmb_{i+1})$,  between successive latent states for a suitable distance $d$ on the simplex \cite{martin-fernandez_measures_1998}. While several choices are possible, for the purpose of this work, we choose the Bhattacharrya-arccos distance \cite{astrom_image_2017}, one that is bounded and differentiable.  We discuss alternative distances in Appendix~ \ref{sec:Appendix:SimplexDistances}. Thus, this regularizer penalizes the total length of the latent state trajectory on the simplex according to  
\begin{align}\label{eq:Rx}
    R_x(\xmb_{1:N}) &=  \sum_{i=1}^{N-1} d^2(\xmb_i, \xmb_{i+1}) \nonumber \\
                    &=\sum_{i=1}^{N-1} \left( \arccos \left(\sum_{k=1}^K \sqrt{\xmb_{i}[k]\xmb_{i+1}[k]}\right) \right).
\end{align}
With the above choice for regularizing the latent state sequence, our choice for regularization of the pure state distributions is motivated by Eq.~\eqref{eq:UniquenessX} and the inverse-scaling nature of barycentric non-uniqueness. Considering our non-uniqueness construction in Eq.~\eqref{eq:UniquenessX}, if we set the reference point to be the current value of the latent state $\xmb_0 = \xmb_t$, and constructed point as the next state $\bar{\xmb}_B = \xmb_{t+1}$, Eq.~\eqref{eq:UniquenessX} takes the form $\xmb_{t+1}= \xmb_t + \frac{1}{\alpha}\dmb_t$ where $\dmb_t$ is a vector in the simplex along which we move the latent state. Since $\dmb_t$ is essentially arbitrary, taking $\alpha>1$ will encourage small changes in $\xmb_t$ as required by Eq.~\eqref{eq:Rx}.  Referring to Fig.~\ref{fig:Uniqueness}, we see that in this $\alpha>1$ regime, barycentric non-uniqueness manifests in the divergence of the quantile functions; i.e., motion along the blue line segments. That this in fact can occur is verified in our experiments in Sec.~\ref{sec:SimulatedInverseScaling}. Now, for time series that reside in each pure state at some period in time, the observed data during those periods will be representative of each of the pure state distributions. Therefore, having estimated pure state distributions diverge is undesirable if we want to accurately learn these quantities. To counteract this diverging behavior, we propose a regularizer in the space of quantile functions that penalizes the sum of squared Wasserstein distances of the pure state quantile functions $\QuantP_{q_k}$ from a reference quantile function $\QuantP_0$. Here, we choose $\QuantP_0$ be the quantile function of $\meas_0 = B(\xmb_0, \meas_{q_{1:K}})$ where $\xmb_0 = \frac{1}{K}\Onemb_K$ (with $\Onemb_K$ the length $K$ vector of all ones): 
\begin{align}\label{eq:Rq}
    &R_q(\meas_{q_{1:K}}) = \sum_{k=1:K} \mathcal{W}_2^2 \left(\meas_{q_k}, B\left(\frac{1}{K}\Onemb_K, \meas_{q_{1:K}}\right)\right)  \nonumber\\
    & \hspace{5mm}= \sum_{k=1:K} \int_0^1 \left(\QuantP_{q_k}(\xi) - \frac{1}{K} \sum_{j=1:K} \QuantP_{q_j}(\xi) \right)^2 d\xi,
\end{align}
The two regularizers just discussed are designed to work in tandem. Here, $R_x$ ensures that the latent state evolves gradually over the simplex while $R_q$ ensures that the pure state quantile functions do not diverge in the ways predicted by the inverse-scaling analysis in Sec.~\ref{sec:Uniqueness}. Through our simulations in Sec.~\ref{sec:SimulatedExperiments}, we demonstrate how by appropriately balancing the regularization weights one can reliably estimate the DWM model parameters.

\subsection{Discrete Quantile Approximation}\label{sec:DQA}
\begin{figure}[h]
  \centering
     \includegraphics[width=0.6\textwidth]{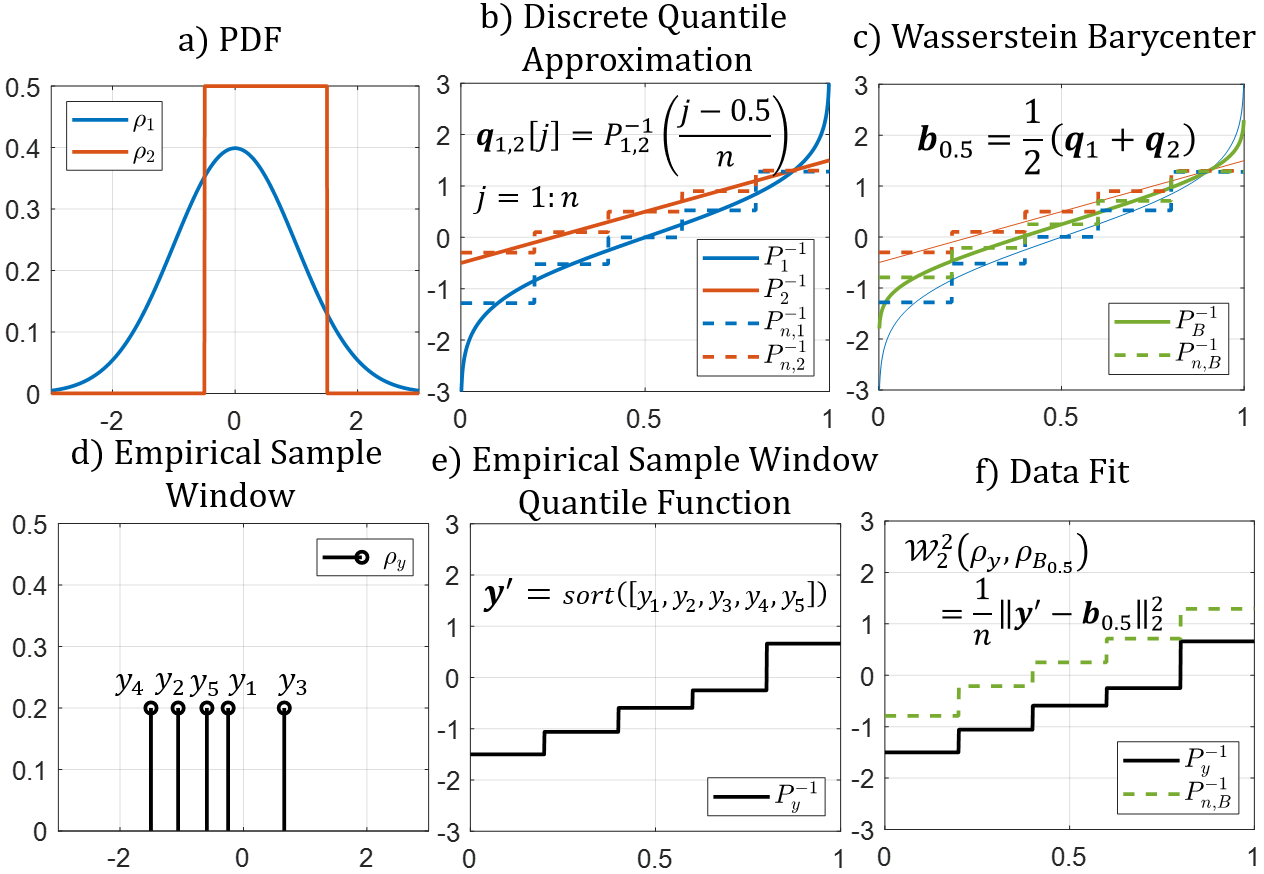}
   \caption{\textbf{Discrete quantile representation for pure states, Wasserstein barycenter, and empirical distribution function.} (a) PDF of $\meas_1, \meas_2$. (b) Respective quantile function and $n$-DQA where $n=5$ with corresponding $n$-DQV $\qmb_{1:2}$.  (c) Quantile function of the Wasserstein barycenter $\meas_{B_{t^*}}=B([0.5,0.5]^\TP, \meas_{1:2})$ and its $n$-DQV, $\bmb_{t^*}$, a weighted average of $\qmb_1$ and $\qmb_2$. (d) Distribution according to Eq.~\eqref{eq:empirical} corresponding to sample window with $n=5$ points. (e), corresponding empirical quantile function $\QuantP_y$ where $\ymb'$ is the sorted vector of samples. (f) Since $\meas_{B_{t^*}}$ and $\QuantP_y$ are monotone step functions sharing the same set of discontinuities, from Eq.~\eqref{eq:1dWass} $\mathcal{W}_2^2(\meas_y, \meas_{n,B})=\frac{1}{n}\norm{\ymb'-\bmb_{0.5}}_2^2$}  \label{fig:DiscreteWass}
\end{figure}

As estimation of the infinite dimensional quantile functions $\QuantP_{q_{1:K}}$ in \eqref{eq:Loss} requires a finite dimensional approximation we introduce the following two quantities:
\begin{definition} \textbf{[n-DQA]}
Given a quantile function $\QuantP: [0, 1] \rightarrow \mathbb{R}$, the \textit{$n$-point discrete quantile approximation} (n-DQA) is a monotone step function $P_n^{-1}(\xi) = \QuantP\left(\frac{\lceil \xi n \rceil - 0.5}{n}\right)$ that is obtained by sampling the $\{\frac{0.5}{n}, \frac{1.5}{n}, ..., \frac{n-0.5}{n}\}$-th quantiles of  $\QuantP$. 
\end{definition}

\begin{definition} \textbf{[n-DQV]}
The \textit{$n$-point discrete quantile vector} (n-DQV) $\qmb \in \realR^n$ is comprised of the sampled quantiles from an $n$-point DQA: $\qmb[j] = \QuantP\left(\frac{j-0.5}{n}\right), j = 1, 2, \cdots, n$.
\end{definition} 

The first two plots in the top row of Fig.~\ref{fig:DiscreteWass} illustrate these definitions. For a univariate distribution, the $n$-DQA approximates the quantile function with a monotone step function whose constant values are sampled from the quantile function on a uniform interval. Since the quantile function is monotone, these sampled quantiles and consequently the $n$-DQV is sorted in ascending order.

With this, we approximate the quantile functions of the pure state distributions $\QuantP_{q_{1:K}}$ with their respective $n$-DQA $\QuantP_{n, q_{1:K}}$, parameterizing them according to their $n$-DQV, which are denoted as $\qmb_{1:K}$. Learning these $n$-DQVs amounts to estimating the $\{\frac{0.5}{n}, \frac{1.5}{n}, \dots, \frac{n-0.5}{n}\}$-quantiles of each pure state distribution. We also use this discrete quantile approach to estimate the quantile functions of the model Wasserstein barycenter $\meas_{B_i}=B(\xmb_i, \meas_{q_{1:K}})$. From Eq.~\eqref{eq:1dWassBary} we see that the $\xi$-th quantile of the Wasserstein barycenter is a weighted combination of the $\xi$-th quantiles of $\meas_{q_{1:K}}$, with barycentric weight $\xmb_i$. Since the $\qmb_{1:K}$ samples the quantile function of each pure state distribution at the same quantile values, the $n$-DQV of $\meas_{B_i}$ is
\begin{align} \label{eq:1dWassBaryDQV}
    \bmb_i =& \sum_{k=1}^K \xmb_i[k] \qmb_k.
\end{align}
We denote corresponding $n$-DQA as $\QuantP_{n,B_i}$ and the corresponding distribution as $\meas_{n,B_i}$.

Using this $n$-DQV representation, by intentionally choosing $n$, the discretization level for the $n$-DQV, to be equal to the size of the sample window, we are able to pose a least-squares cost that approximates the data fit term in Eq.~\eqref{eq:Loss}. To see this, we start by noting that all $n$-DQAs including $\QuantP_{n, q_{k}}, \QuantP_{n,B_i}$ are monotone step functions on $[0,1]$ with discontinuities at $\{\frac{1}{n}, \frac{2}{n}, ..., \frac{n-1}{n}\}$. Additionally, shown in the bottom row of Fig.~\ref{fig:DiscreteWass}, the distributions $\meas_{y_i}$ corresponding to sample windows $\ymb_i$ are discrete distributions comprised of Dirac-delta measures supported on a set of $n$ points with uniform weights (Eq.~\eqref{eq:empirical}). By setting this window size $n$ to be the same as the level of discretization used for the $n$-DQAs that approximate the pure state quantile functions, the quantile functions $\QuantP_{y_i}$ corresponding to $\meas_{y_i}$ will have the same monotone piecewise constant structure with the same set of discontinuities as the $n$-DQA (bottom right of Fig. \ref{fig:DiscreteWass}). With $\ymb'_i$ being the vector obtained by sorting the elements of $\ymb_i$ in increasing order, the Wasserstein distance between the $\meas_{y_i}$ and $\meas_{n,B_i}$ is simply
\begin{align}\label{eq:1dDiscreteWass}
    \mathcal{W}_2^2(\meas_{y_i},\meas_{n,B_i}) =&  \frac{1}{n}\norm{\ymb'_i - \bmb_i}_2^2,
\end{align}
where $n$ is the size of the sample window corresponding to $\meas_{y_i}$ and hence the length of $\ymb'_i$, as well as the level of discretization  used for the $n$-DQA of the pure state distributions, and hence the length of $\bmb_i$. 

\subsection{Model Estimation and Algorithm} \label{sec:ProblemStatement}
Using this discrete quantile parameterization for the pure state distributions allows us to pose an approximation to the variational objective function in Eq.~\eqref{eq:Loss} for the DWB model as a constrained nonlinear least squares problem.
Let $\Qmb=[\qmb_1, ..., \qmb_K] \in \realR^{n \times K}$ denote the matrix whose columns correspond to the $n$-DQV of each of the pure states, $\Ymb = [ \ymb'_1, ..., \ymb'_\T] \in \realR^{n \times \T}$ the matrix whose columns correspond the sorted sample windows from the observed time series and $\Xmb = [\xmb_1, ..., \xmb_\T] \in \realR^{K \times \T}$ the matrix whose columns correspond to the latent state vectors across time. Pulling together the regularizers discussed in Sec.~\ref{sec:Regularization}, we pose the following constrained optimization problem,
\begin{align}
    &\hat{\Qmb},\hat{\Xmb} = \argmin_{\Qmb, \Xmb}  F( \Qmb, \Xmb) \nonumber \\
    & = \argmin_{\Qmb, \Xmb} \underbrace{\frac{1}{n}\norm{\Ymb - \Qmb \Xmb }_{F}^2}_{\text{Data fit}}
            + \lambda_{x} \underbrace{\sum_{i=1}^{\T-1} d^2(\Xmb[:,i+1], \Xmb[:,i])}_{\text{Eq.}~\eqref{eq:Rx}} \nonumber \\
            &\hspace{10mm}+ \lambda_q \underbrace{\frac{\T}{n}\norm{\Qmb \left( \Imb - \frac{1}{K}\Onemb_K \Onemb_K^\TP \right) }_F^2}_{\text{Eq.}~\eqref{eq:Rq}} \label{eq:LossFuncCVX}
\end{align}
subject to:
\begin{align}
    &\Qmb[j+1,k] - \Qmb[j,k] \geq 0, \hspace{3mm}\begin{array}{l} k=1,\dots,K, \\  j=1,\dots,(n-1) \end{array}  \label{eq:ConstraintQ1}\\
    &\sum_{k=1}^K \Xmb[k,i] = 1  \hspace{20mm}  i=1:\T  \label{eq:ConstraintX1}\\
    &\Xmb[k,i] \geq 0  \hspace{24mm}\begin{array}{l} k=1,\dots,K \\ i=1,\dots,\T \end{array} \label{eq:ConstraintX2}  
\end{align}

As detailed in Alg.~\ref{alg:1dDWB}, we minimize Eq.~\eqref{eq:LossFuncCVX} using a \rev{block} coordinate descent approach \rev{with two blocks}, alternating between optimizing for $\Qmb$ and $\Xmb$ while holding the other fixed. For both constrained optimization problems, we utilize the \textit{sequential least squares programming} (SLSQP) optimizer \cite{kraft_software_1988} with python's \texttt{scipy} library \cite{virtanen_scipy_2020}. 

The learned parameters of $\hat{\Qmb}$ and $\hat{\Xmb}$ both have \rev{closed, non-empty, and} convex constraints given by Eqs.~\eqref{eq:ConstraintQ1}-\eqref{eq:ConstraintX2}. The $n$-DQV, and thus the columns of $\Qmb$ must be sorted in ascending order (Eq.~\eqref{eq:ConstraintQ1}). The barycentric weights, and thus the columns of $\Xmb$, are constrained to the set of positive matrices with rows that sum to one (Eqs.~\eqref{eq:ConstraintX1}, \eqref{eq:ConstraintX2}). \rev{Thus, by \cite{grippo_convergence_2000}, every limit point of an alternating block coordinate descent approach with two blocks is guaranteed to be a critical point. We show empirical convergence of this algorithm to a limit point in Fig.~\ref{fig:Convergence}}

Regarding the initialization of the parameters $\hat{\Qmb}$ and $\hat{\Xmb}$, we initialize the pure state distributions based on clustering the observation sample windows. Starting in line \ref{alg:line:cluster} in Alg.~\ref{alg:1dDWB}, we compute the similarity graph using the exponential of the negative square Wasserstein distance between the distributions estimated from any two sample windows and use spectral clustering from Python's \texttt{sklearn} \cite{buitinck_api_2013} to learn $K$ clusters. Denoting $\mathcal{C}_k$ as the index set of sample windows that belong to cluster $k$, in line \ref{alg:line:init} we initialize the $n$-DQV of each pure state to be, $\qmb_k = \frac{1}{\abs{\mathcal{C}_k}} \sum_{i \in \mathcal{C}_k} \ymb'_{i}$.
We initialize the latent state $\hat{\Xmb}[t,k]=\frac{1}{K} \forall i=1,\dots,\T, \, k=1,\dots,K$, to be at the centroid of the simplex for all points in time.

\begin{algorithm}[h]
    \SetAlgoLined
     \SetKwInOut{Input}{Input}
     \SetKwInOut{HyperParams}{HyperParams}
     \SetKwInOut{Output}{Output}
     \SetKwRepeat{Do}{do}{while}
    
    \textbf{Input:} \\ 
        $y_1,...,y_t, ..., y_T$: Univariate time series \\ 
        $t_1, ..., t_i, ..., t_\T$: Starting indices for sample windows\\
        $K$: Number of pure states \vspace{1mm} \\ 
    \textbf{Hyperparameters:} \\ 
        $n$: Sample window size \hspace{2mm} $\eta$: Convergence threshold\\
        $\lambda_x, \lambda_q$: Regularization weights to define $F$ (Eq.~\eqref{eq:LossFuncCVX}) \vspace{2mm}\\
        
    \textbf{Output:} \\ 
    $\hat{\Qmb} \in \realR^{n \times K}$: Stacked pure state DQVs \\  
    $\hat{\Xmb} \in \realR^{K \times \T}$ Stacked barycentric weights \vspace{2mm} \\
    
    \For{$i=1:\T$} {  
        $\ymb'_i = \textit{sort}(y_{t_i}, ..., y_{{t_i}+n})$ \tcp*{sorted windows}
    }
    \For(\tcp*[f]{Window affinity matrix}){ \label{alg:line:cluster} $i=1:\T$} { 
        \For{$j=1:\T$} {
            $\mathbf{A}[i,j] = \frac{1}{n}\norm{\ymb'_i - \ymb'_j}_2^2$ 
        }
    }
    $c_{1:\T} = \textit{SpectralClustering}(K, \exp(-\mathbf{A}))$ \tcp*{Cluster sample windows}
    \For{$k=1,...,K$} {  
        $\mathcal{C}_k = \{i: c_i=k\}$ \\
        $\qmb_k = \frac{1}{\abs{\mathcal{C}_k}}\sum_{i \in \mathcal{C}_k} \ymb'_{i}$ \tcp*{Init. $n$-DQV} \label{alg:line:init}
    }
    $\Ymb = \left[ \ymb'_1, ..., \ymb'_\T \right]$ \tcp*{Stacked windows}
    $\hat{\Xmb}^{(0)} = \frac{1}{K}\Onemb_{K} \Onemb_{\T}^\TP$ \tcp*{Initialize $\Xmb$}
    $\hat{\Qmb}^{(0)} = \left[ \qmb_1, ..., \qmb_K \right]$  \tcp*{Stacked $n$-DQVs}

     \Do {\text{$F \left(\hat{\Qmb}^{(i)}, \hat{\Xmb}^{(i)}\right)- F\left( \hat{\Qmb}^{(i+1)}, \hat{\Xmb}^{(i+1)} \right) > \eta$}} {
        $\hat{\Xmb}^{(i+1)} = \argmin_{\Xmb} F\left(\Qmb = \hat{\Qmb}^{(i)}, \Xmb = \hat{\Xmb}^{(i)}\right)$ \tcp*{SLSQP}
        $\hat{\Qmb}^{(i+1)} = \argmin_{\Qmb} F\left(\Qmb = \hat{\Qmb}^{(i)}, \Xmb = \hat{\Xmb}^{(i+1)}\right)$ \hspace{-1mm}\tcp*{SLSQP}
     }
\caption{Non-parametric and Regularized DWB }\label{alg:1dDWB}
\end{algorithm}

\section{Model Evaluation} \label{sec:Evaluation}
Analyzing simulated and real world human activity data where the system gradually evolves among its pure states, we empirically demonstrate (1) how the proposed regularizers allow the DWB model to accurately recover the system parameters taking into account the inverse-scaling relationship between the model parameters, (2) the impact of the window size on the accuracy of estimating the system parameters of a dynamically evolving time series, and (3) how our non-parametric formulation of the DWB problem is able to accurately learn the pure state distributions. 

\subsection{Simulated Experiments} \label{sec:SimulatedExperiments}
\begin{figure}
 \centering
    \includegraphics[width=0.489\textwidth]{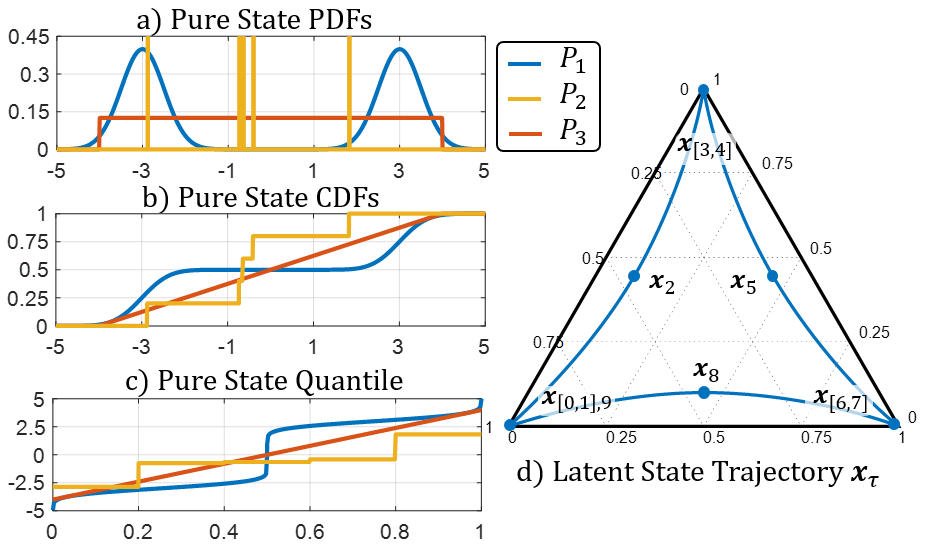}
  \caption{\textbf{Simulated pure state distributions and latent state process.} (a) PDF, (b) CDF, and (c) quantile function for the pure state distributions used in the simulated experiments. (d) Simulated latent state trajectory as the system transitions among its three pure states. }  \label{fig:PureStates}
\end{figure}

Our approach proposes regularizers to the DWB model with the assumption that the latent state of the system transitions gradually among its pure states. Here, we consider a simulated system whose parameters reflect these properties, denoting the ground truth pure state distributions and latent states as $\meas_{q_{1:K}}$, $\xmb_{1:T}$ and the model estimated parameters as $\hat{\meas}_{q_{1:K}}$, $\hat{\xmb}_{1:T}$. 

Consider a system that consists of $K=3$ pure states with ground truth distributions,
\begin{align}
    \meas_{q_1} =& 0.5 \,\mathcal{N}(3, 0.25) + 0.25 \, \mathcal{N}(-3, 0.25)  \\
    \meas_{q_2} =& \mathcal{U}[-4, 4] \nonumber \\
    \meas_{q_3} =& \frac{1}{5}\sum_{i=1}^5 \mathcal{N}\left(\qmb[i], 1e^{-8} \right) \nonumber \\
    & \hspace{5mm}\qmb = \left[ -2.88, -0.74, -0.64, -0.41, 1.82 \right]. \nonumber
\end{align} 
Here, $\mathcal{U}[a,b]$ denotes a uniform distribution on the interval $[a,b]$. We show the PDF in Fig.~\ref{fig:PureStates}a, CDF in Fig.~\ref{fig:PureStates}b, and quantile functions in Fig.~\ref{fig:PureStates}c for each of these distributions 

Now let us consider a time series that transitions among these three pure states following the trajectory outlined in Fig.~\ref{fig:PureStates}d. The continuous-time latent state $\xmb_\tau$ alternates between pausing at each pure state for $1$ second, and then transitioning to another pure state for $2$ seconds moving from $\meas_{1}\ldots \meas_{2} \ldots \meas_{3} \ldots \meas_{1}$ over the course of continuous time $\tau=[0,9]$ seconds. To emulate the simulated setup in Sec.~\ref{sec:Windowing} of varying the per-sample change in distribution, we vary $r$ (Hz), the rate at which we sample this continuous-time sequence to generate the ground truth latent state $\xmb_t$ for $t=1,\dots,T$. Time series are independently sampled $y_t \sim \meas_{B_t} = B(\xmb_t, \meas_{q_{1:3}})$ by uniformly sampling a quantile $\xi \in [0,1]$ and evaluating the quantile function $\QuantP_{B_t}(\xi) = \sum_{k=1}^3 \xmb_t[k] \QuantP_{q_k}(\xi)$. 

From this time series, we generate a sequence of sample windows using sliding window\cite{aminikhanghahi_survey_2017}, spacing out the windows on a constant interval, $t_{i+1} = t_i +\delta$. For these experiments, we choose to set $\delta=n$, which partitions the time series into a sequence of $\T = \lfloor \frac{T}{n} \rfloor$ disjoint windows of size $n$.

With knowledge of ground truth, we can assess our model using the average distance of our learned parameters to these ground truth values according to,  
\begin{align} \label{eq:EvalGT}
    e_q = \frac{1}{K} \sum_{k=1}^K \mathcal{W}_2^2 \left(\meas_{q_k}, \hat{\meas}_{q_k} \right) \hspace{4mm}
    e_x = \frac{1}{\T} \sum_{t=1}^\T  \norm {\xmb_t - \hat{\xmb}_t}_2^2.
\end{align}
Since our model is unsupervised, there is no guarantee that the indexing of the learned pure states will match that of the ground truth. Therefore, when assessing our model, we assume that learned pure states (and subsequently the latent state vector) are reordered in a manner that minimizes Eq.~\eqref{eq:EvalGT}.

\subsubsection{Regularization and Inverse-Scaling} \label{sec:SimulatedInverseScaling}
\begin{figure*}
 \centering
    \includegraphics[width=1.0\textwidth]{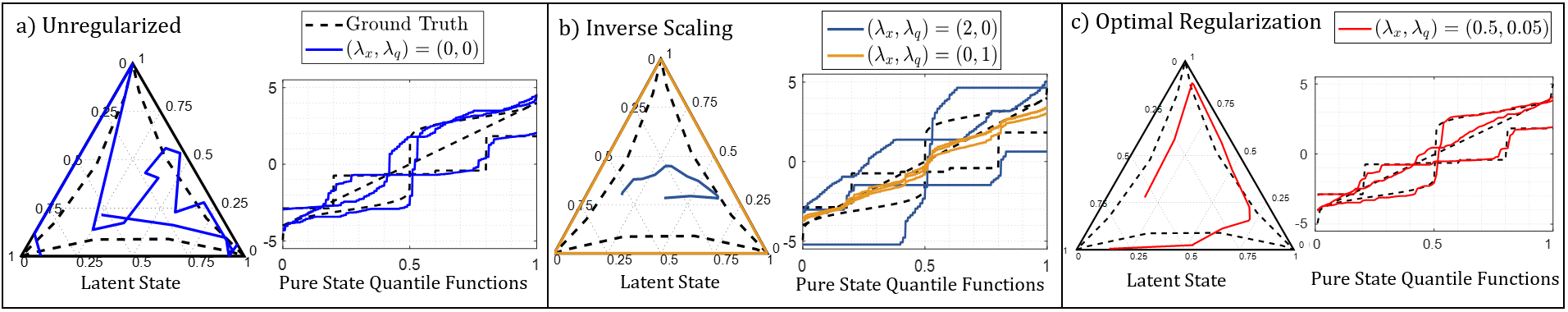}
  \caption{\textbf{Impact of regularization and the inverse-scaling relationship}. (a) Without regularization, though the ground truth latent state evolves gradually, the learned latent state varies over the simplex. (b) Introducing just the latent state regularizer (blue) imposes some level of smoothness to the latent state, but consequently results in the pure state quantile functions diverging from their ground truth values. Conversely, including just the pure state regularizer (orange) causes the pure state quantile functions to move closer together and the latent states to move towards the boundary of the simplex. (c) Using regularization weights that minimize $e_x+e_q$ balance the effects of these two regularizers to accurately recover the gradually evolving latent state sequence and the pure states quantile functions of the time series. Plots are shown for $r=200, n=100$.}  \label{fig:CounteractQX}
\end{figure*}

We demonstrate the inverse scaling relationship between the model parameters through the interaction between the latent state and pure state regularizers. As discussed in Sec.~\ref{sec:Windowing}, the difficulty in accurately estimating the model parameters stems from the problem of estimating dynamically evolving data distribution with a window of finite length. In Fig.~\ref{fig:CounteractQX}, we see that in the absence of any regularization, the learned latent state can vary over the simplex, even when the ground truth latent state is stationary or gradually evolving.
The blue lines of Fig.~\ref{fig:CounteractQX}b show that introducing only the latent state regularizer that penalizes $d^2(\xmb_t, \xmb_{t+1})$ imposes some level of smoothness on the latent state. However, as discussed in Sec.~\ref{sec:Regularization} this regularizer has an additional effect that is similar to the $\alpha>1$ regime corresponding to the blue lines in Fig.~\ref{fig:Uniqueness} where the latent states trajectory contracts towards a point on the simplex. In this case, as specified by the inverse-scaling relationship, seen from the blue quantile plots in Fig.~\ref{fig:CounteractQX}b, the pure state quantile functions diverge from the ground truth values. 
On the other hand, as seen in orange lines of Fig.~\ref{fig:CounteractQX}b, having only the pure state regularizer causes the pure state quantile functions to be pulled closer together where by the inverse-scaling relationship, the latent states move away from the centroid of the simplex towards the boundary. 

Only through the combination of these regularization terms can we recover the ground truth parameters of this simulated system that gradually evolves among its pure states. Using our knowledge of ground truth we perform a grid search by varying the regularization weights $\lambda_x, \lambda_q = \{1e^{-4}, 2e^{-4}, 5e^{-4}, 1e^{-3},..., 1e^{-1} \}$ and picking the pair that minimizes the total ground truth error $e_x + e_q$. As seen from Fig.~\ref{fig:CounteractQX}c, the DWB model corresponding to these optimal regularization weights more accurately recover the ground truth parameters of this simulated system. In Sec.~\ref{sec:RealWorldResults}, we perform parameter selection in the absence of ground truth using L-surfaces \cite{brooks_inverse_1999}.

\subsubsection{Impact of Window Size} In Sec.~\ref{sec:Windowing}, we highlighted the effect of window size and the rate of change in the data distribution on the accuracy of estimating a dynamically evolving distribution using a window of samples. Here, we seek to understand how these factors impact the accuracy of the learned DWB parameters. Using the aforementioned setup in Sec.~V-A, we simulate two systems with different sampling rates $r\in\{150, 200\}\, \text{Hz}$, generating 500 time series per value of $r$. We then run our DWB model varying the window size between $n=50,100,...,400, 410, ...,500$. We find the optimal regularization weights by performing a grid search over the range of values as Sec.~V-A-1. For each value of $n$ and $r$, we set $\lambda_x, \lambda_q$ to be the average of the results of this grid search performed for 5 randomly selected time series.

\begin{figure}
 \centering
    \includegraphics[width=0.489\textwidth]{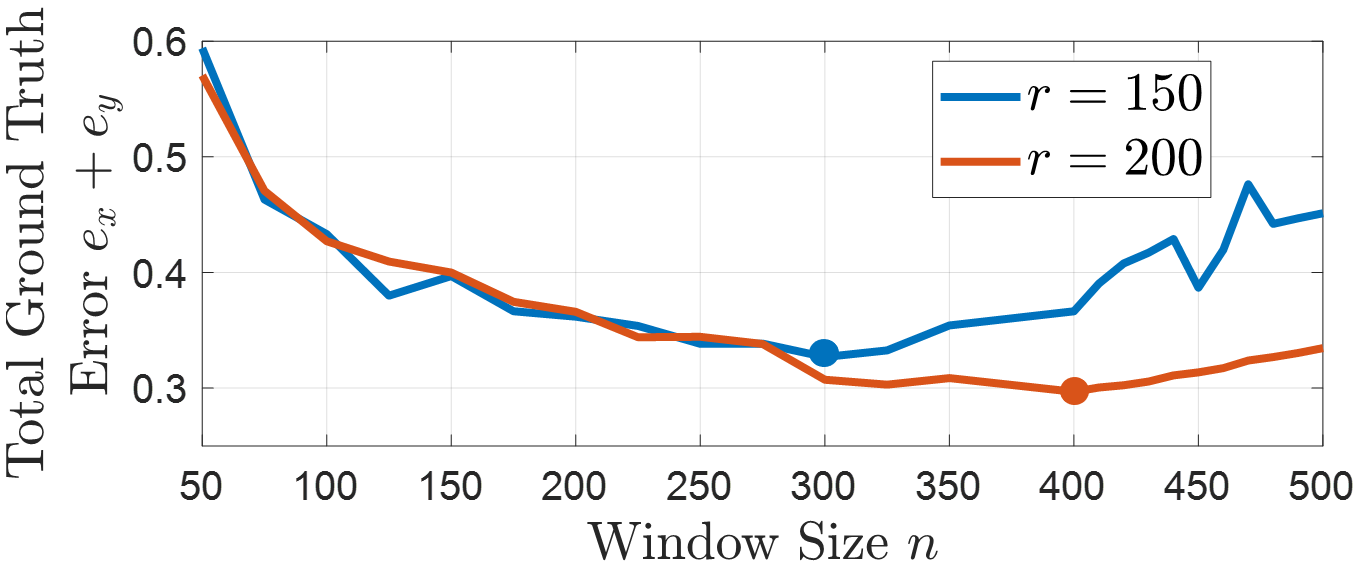}
  \caption{\textbf{Impact of window size on simulated model accuracy}. Ground truth error averaged over 500 generated time series as a function of window size $n$ for sampling rates of $r={150, 200}$ Hz. The U-shape plots imply that the factors that impact the ability of small and large windows to estimate a dynamically evolving data distribution similarly impact the ground truth error of the learned DWB model parameters. Additionally, increasing the sampling rate $r$ of the time series, which results in a smaller per-sample change in the data distribution, similarly improves the model accuracy for larger windows and shifts the minimum of the U-curve towards larger windows. \rev{The minimum error (solid dots) for $r=150$ was achieved with a window size of $n=300$, while for $r=200$ the minimum window size was achieved at $n=400$.} }  \label{fig:ErrorVsWindow}
\end{figure}

In this experiment, we remove potential confounding variables across the various window sizes. First, we use the same initialization for the pure states distributions, specifically the one generated from the spectral clustering method discussed in Sec.~\ref{sec:ProblemStatement} for the configuration of $n=200$. Secondly, we ensure that the sequence of sample windows estimates the time series at the same points in time, such that regardless of $r,n$ the windows are centered at $\tau=[1.5, 2.0, ..., 7.5]$. 

The results shown in Fig.~\ref{fig:ErrorVsWindow} imply that the factors discussed in Sec.~\ref{sec:Windowing} that impact the ability of a sample window to accurately estimate a dynamically evolving data distribution similarly affect the ability of the DWB model to accurately learn a system's pure states and latent states from these estimated windows. The U-shape curves in Fig.~\ref{fig:ErrorVsWindow} of the average ground truth error as a function of window size bear strong similarities to the U-shape curves illustrating the window size tradeoff in Fig.~\ref{fig:WindowExperimentResults} where small windows lack the samples to precisely estimate the data distribution and the accuracy of large windows suffer due to the dynamics of the systems. Furthermore, we see the same trend as in Sec.~\ref{sec:Windowing} where decreasing the per-sample change in distribution, shown here by increasing the sampling rate $r$, improves the accuracy for models with larger windows shifting this window size tradeoff and the ``optimal'' window size towards larger windows.

\subsubsection{Convergence}
\rev{Convergence of the optimization process to a limit point is shown in Fig.~\ref{fig:Convergence} for various configurations of the simulated data defined in Sec.~\ref{sec:SimulatedExperiments}.  
\begin{figure}
 \centering
    \includegraphics[width=0.45\textwidth]{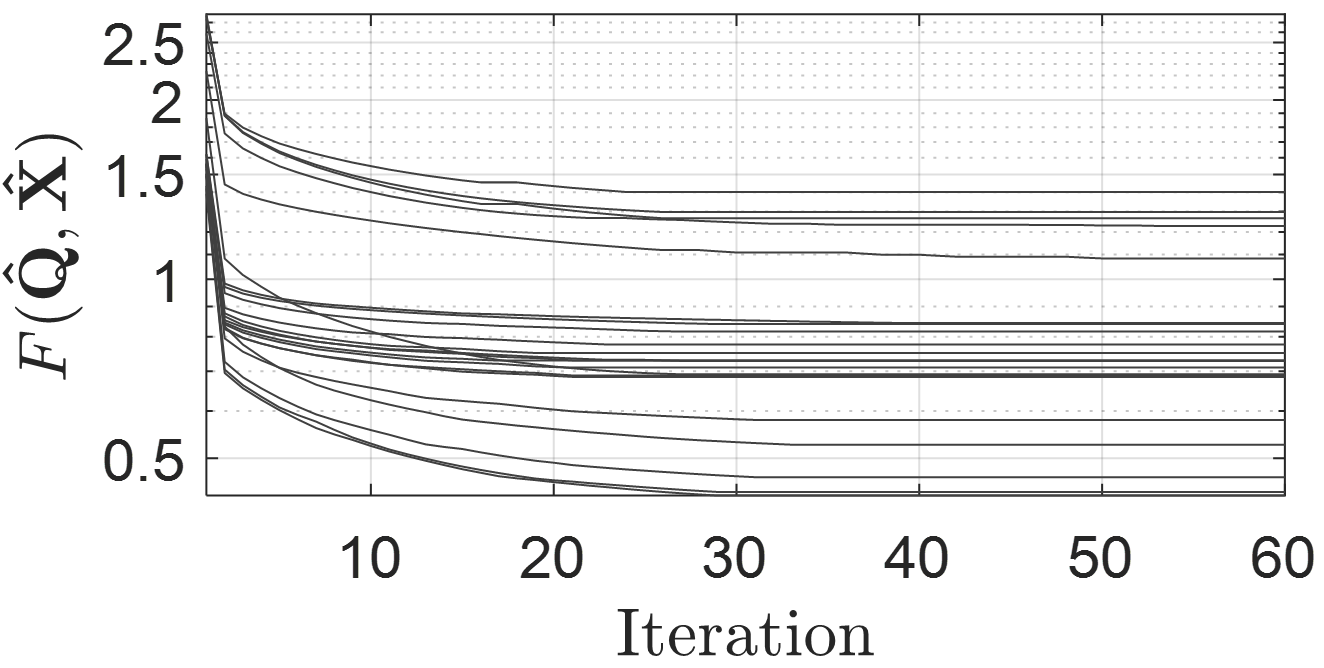}
  \caption{\rev{\textbf{Convergence of simulated experiments:} Plot of the objective function shows convergence to a limit point. Shown for simulated data with $n=400, 410, ..., 500$ and $r=150, 200$.}}  \label{fig:Convergence}
\end{figure}
}
\subsection{Real World Data} \label{sec:RealWorldResults}
In this section, we compare the performance of the non-parametric DWB model to the Gaussian DWB model \cite{cheng_dynamical_2021} on univariate data. To highlight the difference between the Gaussian and non-parametric discrete quantile parameterizations, we use the regularization framework proposed in this paper for both models. We evaluate using our simulated data and two human activity datasets.
\begin{enumerate}
    \item \textit{Beep Test (BT, proprietary):} Subjects run between two points to a metronome with increasing frequency, alternating between two states: running and standing. We use the vertical component (z-axis) of the 3-axis accelerometer, which is the dimension in which the distributions of the two pure states are best differentiated. The sensor is sampled at 100 Hz.

    \item \textit{Microsoft Research Human Activity (MSR, \cite{morris_recofit_2014}):} 126 subjects perform exercises in a gym setting. Exercises vary among subjects covering strength, cardio, cross-fit, and static exercises. Each time series is truncated to five minutes. Discrete labels corresponding to activities are provided, thus we set $K$ to the number of labeled discrete states in the truncated time series (range: $K=2$ to $7$). We use the x-axis of the 3-axis accelerometer sampled at 50 Hz. 
    
    \item \textit{Simulated Data (Sim):} Following the data generating process outlined in Sec.~\ref{sec:SimulatedExperiments}, we simulate 500 time series setting $r=200$.
\end{enumerate}

\textbf{Evaluation:} Since ground truth is not known in the real world setting, we assess performance by considering the model fit to the data,  computed using the data-fit term in Eq.~\eqref{eq:Loss},
\begin{align} \label{eq:ey}
    e_y &= \frac{1}{\T} \sum_{t=1}^\T \mathcal{W}_2^2 \left(\meas_{y_t}, \meas_{B_t} \right)
\end{align}
In the non-parametric model, this distance is computed according to Eq.~\eqref{eq:1dDiscreteWass}. In the Gaussian DWB model where $\meas_{B_t}$ is Gaussian, this distance is computed using a Monte-Carlo method using $1e^5$ IID samples from $\meas_{B_t}$.  
    
\begin{figure}[h]
 \centering
    \includegraphics[width=0.6\textwidth]{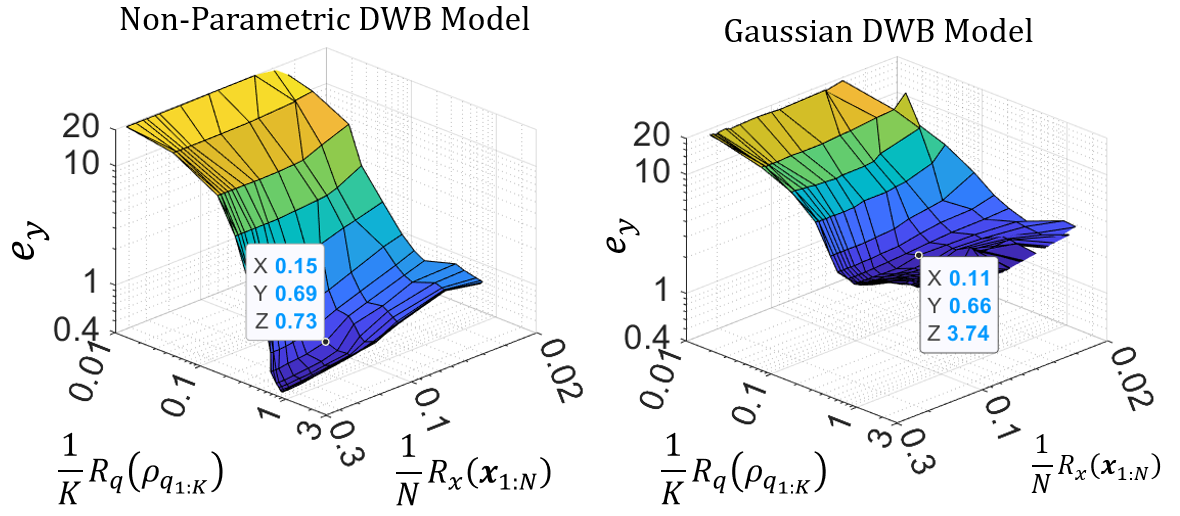}
  \caption{\textbf{L-surface for selection of regularization weights}. L-surface for MSR time series for (left) non-parametric and (right) Gaussian DWB models varying $\lambda_x, \lambda_q$ from $[1e^{-5}, 1]$. Each figure plots the magnitude of the data loss against the magnitude of the two regularization parameters on a log scale. Selecting the regularization weight according to L-surfaces amounts finding the $\lambda_x, \lambda_q$ corresponding to the ''corner`` of the surface plot, the point where further increasing $\lambda_x, \lambda_q$ (which corresponds to moving toward the back corner of the plot, decreasing $\frac{1}{K}R_q(\meas_{q_{1:K}})$ and $\frac{1}{N}R_x(\xmb_{1:T})$) results in a sharp increase in the magnitude of the data loss.}  \label{fig:LSurface}
\end{figure}

\begin{figure*}
 \centering
    \includegraphics[width=1.0\textwidth]{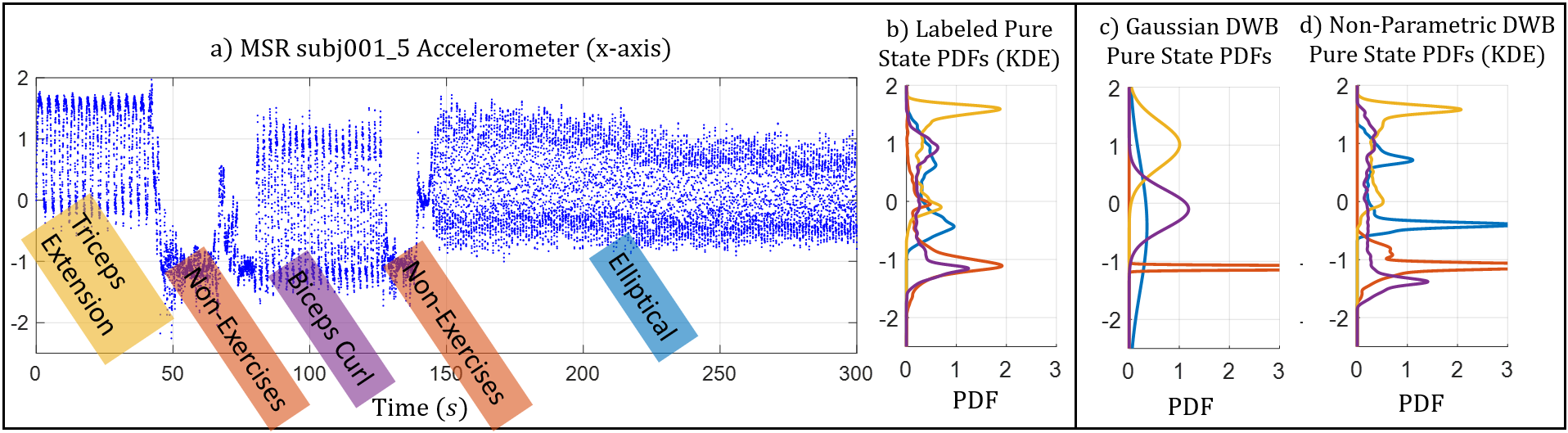}
  \caption{\textbf{Non-parametric vs Gaussian pure states distributions for MSR data}. (a) MSR time series consisting of 4 activities over 5 minutes. (b) Pure state distributions estimated from discrete labeled data converted to a PDF using KDE with a Gaussian kernel ($\sigma=0.04$). (c) Learned pure states for Gaussian model from \cite{cheng_dynamical_2021}. (d) Learned pure states for our proposed non-parametric model. The $n$-DQV $\qmb_k$ for each pure state is converted to a PDF using KDE with the same Gaussian kernel centered on the values of $\qmb_k$. The distribution of the data in each pure state activity (blue: Elliptical, yellow: Tricep Extension, purple: Bicep Curl, Red: Non-exercise) is better captured using the non-parametric approach $e_y=0.73$ compared to the Gaussian $e_y=3.74$.}  \label{fig:MSR}
\end{figure*}
To select regularization weights when ground truth is not available, we draw from the field of inverse problems and use L-surfaces \cite{brooks_inverse_1999}.
Fig.~\ref{fig:LSurface} shows the L-surface for one MSR dataset plotting $e_y$ against the magnitude of the $R_{x}$ and $R_q$ on a log scale while varying $\lambda_x, \lambda_q$ ranging from $1e^{-5}$ to $1$. For each model configuration and dataset, we use this L-surface method to pick a set of regularization parameters based on one representative time series and apply those parameters to the rest of the dataset. These parameters are detailed in  Tab.~\ref{tab:RealWorldConfig}.

Fig.~\ref{fig:MSR} compares the learned pure states of the Gaussian DWB and non-parametric DWB for one example MSR time series. As seen from Fig.~\ref{fig:MSR}b, the distributions of the data corresponding to many of the activities are clearly multi-modal and therefore not Gaussian. Compared to the Gaussian DWB approach shown in Fig.~\ref{fig:MSR}c, the pure states learned using the proposed non-parametric representation shown in Fig.~\ref{fig:MSR}d more closely match the estimated pure states from the data.
This is also reflected quantitatively in Tab.~\ref{tab:RealWorldResults} where the values of $e_y$ show that our non-parametric DWB model better approximates the sample windows of the time series compared to the Gaussian DWB model for each of the evaluated datasets. 

We note that the magnitude of the $e_y$, and thus the results in Tab.~\ref{tab:RealWorldResults} are dependent on the choice of the regularization parameters, which are chosen in a partially subjective manner. However, we can see in the L-surfaces in Fig.~\ref{fig:LSurface} that compared to the Gaussian model the error is significantly lower in the non-parametric model. Therefore any small subjective changes in the choice of regularization parameter would not significantly alter these conclusions.

\begin{table}[]
  \centering
  \begin{tabular}{|c|c|c|c|}
    \hline
       & MSR & BT & Sim \\
    \hline
        $n$ & 250 & 100 & 100 \\
    \hline
        NP $\lambda_x, \lambda_q$ & $5e^{-2}, 5e^{-3}$ & $2e^{-1}, 5e^{-2}$ & $2e^{-1} 1e^{-2}$ \\
    \hline
        Gauss $\lambda_x, \lambda_q$ & $5e^{-2}, 2e^{-3}$ & $2e^{-1}, 2e^{-2}$ & $1e^{-1} 2e^{-3}$ \\
    \hline
    \end{tabular}
    \caption{\textbf{DWB Model configuration}. Window size and regularization weights for non-parametric (NP) and Gaussian (Gauss) DWB models for simulated (Sim) and real world (BT, MSR) datasets.}
    \label{tab:RealWorldConfig}
\end{table}

\begin{table}[]
  \centering
  \begin{tabular}{|c|r|r|r|}
    \hline
        & \multicolumn{1}{c|}{MSR} & \multicolumn{1}{c|}{BT} & \multicolumn{1}{c|}{Sim}  \\
    \hline
        NP & \textbf{1.32} & \textbf{2.07} & \textbf{2.19}  \\
    \hline
        Gauss & 3.76 & 6.00 & 12.33  \\
    \hline
    \end{tabular}
    \caption{\textbf{Quantitative comparison of DWB model}. Average $e_y$ across all time series for simulated (Sim) and real world (BT, MSR) datasets show clear benefits of the non-parametric DWB model \rev{compared to the Gaussian data-generating distribution model used by \cite{cheng_dynamical_2021}.}}
    \label{tab:RealWorldResults}
\end{table}

\section{Conclusions and Future Work} \label{sec:Conclusions}
In this work, we build upon the DWB model in \cite{cheng_dynamical_2021}.  \rev{We present and discuss the inverse scaling relationship which captures the lack of uniqueness in the DWB model.  We also consider }the challenges of estimating the data distribution of a dynamically evolving time series.  We then propose a temporal smoothness regularization framework to simultaneously address both of these challenges.   Finally, we move beyond the Gaussian assumption of \cite{cheng_dynamical_2021} by using a discrete approximation to the pure state quantile function which results in a least-square ``data-fit'' term in the DWB objective function. Using simulated data, we demonstrate how the two proposed regularization terms work together to achieve the desired smoothness in the time evolution of the latent state as well as the impact of window size on the accuracy of the learned model. In the real world setting of human activity analysis, we demonstrate how compared to the original Gaussian DWB model our non-parametric DWB model better characterizes the time-varying data distribution of the time series. 

An important future work is to extend the univariate framework to the general non-parametric multivariate setting. Lack of closed form expressions for barycenters in the non-parametric multivariate setting as well as the ``curse of dimensionality'' in approximating a high-dimensional distribution from samples \cite{weed2019sharp} makes this extension numerically and statistically challenging. However, fast algorithms for computing Wasserstein barycenters \cite{cuturi_fast_2014} can be effectively utilized for low dimensional problems.

Furthermore, in this work we only consider the Wasserstein barycenter to model the change in distribution as a system moves among pure states. An interesting topic for future research is to consider barycenters corresponding to alternative probability distribution distances such as the Sinkhorn divergence \cite{shen_sinkhorn_2020}.

\rev{Our choice to approximate the quantile function of $\meas_{q_k}$ by sampling on a uniform interval combined with the use of a Dirac-delta functions to estimate $\meas_{y_i}$ from windows of samples lead to the finite dimensional least-squares data fit term in Eq.~\eqref{eq:LossFuncCVX}. Future work can consider alternative methods of estimating distributions from samples (e.g kernel-density estimation \cite{harvey_kernel_2012}) and quantile approximation methods (e.g linear or spline decomposition \cite{daehlen_decomposition_1992}) to derive alternative forms of this univariate DWB problem.}

\rev{We empirically show the monotonic convergence of the two-block cyclic descent method used in the paper.  The theory in \cite{grippo_convergence_2000} indicates that we are converging to a critical point of the cost function.  In future work, we can seek stronger guarantees of converging to a second-order stationary point using a proximal point block coordinate descent algorithm \cite{li_alternating_2019}.}
 
Finally, in this work, we also explore the issue of approximating the data distribution of a dynamically evolving system from a window of \rev{independent, non-identically distributed random variables. We demonstrate the tradeoff between the errors associated with large and small windows using} simulated data to illustrate the effects of system parameters that drive the per-sample change in the distribution on the accuracy of this window estimate. Future work may consider an analytical approach to this problem using the relationship between quantiles and order statistics \cite{david_order_2004} \cite{sen_note_1970} and by bounding the maximum change in distribution over the length of the window.

\bibliographystyle{./bibliography/IEEEtran}
\bibliography{./bibliography/IEEEabrv,./bibliography/IEEEexample}


\begin{IEEEbiographynophoto}{}
\end{IEEEbiographynophoto}

\appendices
\section{Extensions of DWB Non-Uniqueness} \label{sec:Appendix:Uniqueness}
\begin{figure}[!ht]
 \centering
    \includegraphics[width=0.6 \textwidth]{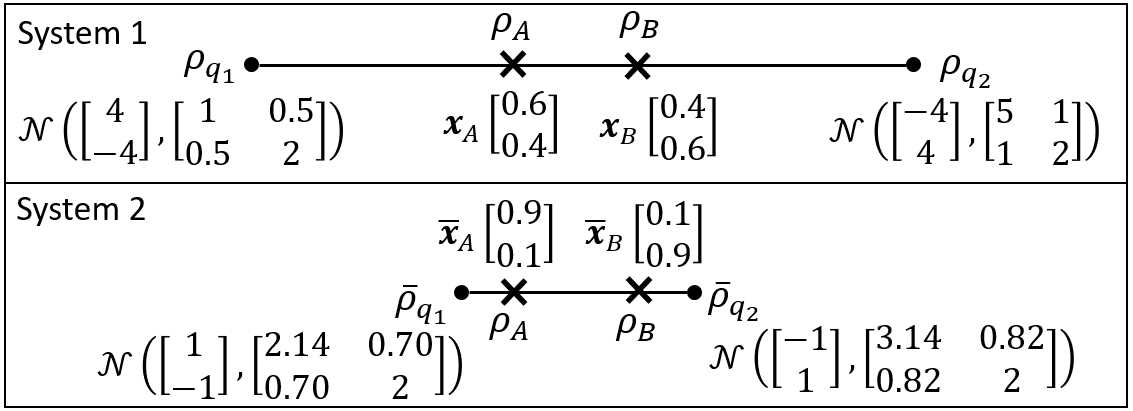}
  \caption{\textbf{Non-uniqueness of Wasserstein barycenter parameters for multivariate Gaussians}. Both systems of pure state distributions and barycentric weights result in the same barycenters, $\meas_A=B(\xmb_A,\meas_{q_{1:2}})=B(\bar{\xmb}_A,\bar{\meas}_{q_{1:2}})$ and $\meas_B=B(\xmb_B,\meas_{q_{1:2}})=B(\bar{\xmb}_B,\bar{\meas}_{q_{1:2}})$. Consistent with the inverse-scaling relationship discussed in Sec.~\ref{sec:Uniqueness}, compared to system 1, in system 2 the Wasserstein distance between $\bar{\meas}_{q_1}$ and $\bar{\meas}_{q_2}$ is diminished while the distance between $\bar{\xmb}_A$ and $\bar{\xmb}_B$ is increased.\label{fig:GaussianUniqueness}}
\end{figure}

While the discussion of this paper pertains mainly to the univariate case, the issue of uniqueness discussed in Sec.~\ref{sec:DWB_Model} also exists in the multivariate case. Consider the two dimensional systems specified in Fig.~\ref{fig:GaussianUniqueness} of systems with two Gaussian pure states. For both systems, $\meas_A=B(\xmb_A,\meas_{q_{1:2}})=B(\bar{\xmb}_A,\bar{\meas}_{q_{1:2}})$ and $\meas_B=B(\xmb_B,\meas_{q_{1:2}})=B(\bar{\xmb}_B,\bar{\meas}_{q_{1:2}})$.

This multivariate example also exhibits the inverse scaling relationship discussed in Sec.~\ref{sec:Uniqueness}. The Wasserstein distance between the pure states in system 1 ($\meas_{q_1}, \meas_{q_2}$) is \textit{larger} than that of system 2 ($\bar{\meas}_{q_1},\bar{\meas}_{q_2}$), however, the resulting distance between the latent states corresponding to $\meas_A, \meas_B$ in system 1 ($\xmb_A, \xmb_B$) is \textit{smaller} than that of system 2 ($\bar{\xmb_A}, \bar{\xmb_B}$).

Furthermore, we can create an example where the construction specified in Sec.~\ref{sec:Uniqueness} does not result in multiple possible values of $\xmb_B$ and $\QuantP_{q_{1:K}}$ and thus cannot be used to demonstrate the non-uniquness of the Wasserstein barycenter parameters. 

Let $\meas_{q_1}=\delta_0$ and $\meas_{q_2}= U[0,1]$, where $\QuantP_{q_1}(\xi)=0$ and $\QuantP_{q_2}(\xi) = \xi$ for $\xi \in [0,1]$. Then the Wasserstein barycenter $\meas_t = B(\xmb_t, \meas_{1:2})$ for any barycentric weight $\xmb_t = [(1-t), t]^\TP$ will have distribution $U[0,t]$ and quantile function $\QuantP_t(\xi) = t\xi$. Per the construction in Sec.~\ref{sec:Uniqueness}, let us pick $\xmb_B = [1,0]$ to be at a vertex, thus $\QuantP_B(\xi) = 0$ and let $\xmb_0 =[0.5, 0.5]^\TP$, thus in $\QuantP_0(\xi)=0.5\xi$. According to Eq.~\eqref{eq:UniquenessX}, $\bar{\xmb}_B$ falls of the simplex for any $\alpha<1$ thus making $\alpha_0=1$. Similarly, according to Eq.~\eqref{eq:UniquenessP}, $\bar{P}^{-1}_B(\xi) = (1-\alpha)t\xi$, which breaks the monotonically increasing constraint of quantile functions for any $\alpha >1 $, thus restricting $\alpha_m=1$. Thus, according to the construction in Sec.~\ref{sec:Uniqueness}, the set $[\alpha_0, \alpha_m]=1$.

Although this example breaks the argument of the non-uniqueness of the Wasserstein barycenter parameters for this specific construction used to highlight the inverse-scaling relationship, it does not mean that the parameters of the Wasserstein barycenter are indeed unique. For example, since $\xmb_B$ is taken to be at a vertex $[1,0]^\TP$ resulting in $\meas_B = \meas_{q_1}$, any valid distribution can be chosen for $\meas_{q_2}$ and still have $\meas_B = B([1,0]^\TP, \meas_{q_{1:2}})$.

\section{Alternative Simplex Distances} \label{sec:Appendix:SimplexDistances}

    \begin{figure}[h]
    \centering
        \includegraphics[width=0.8\columnwidth]{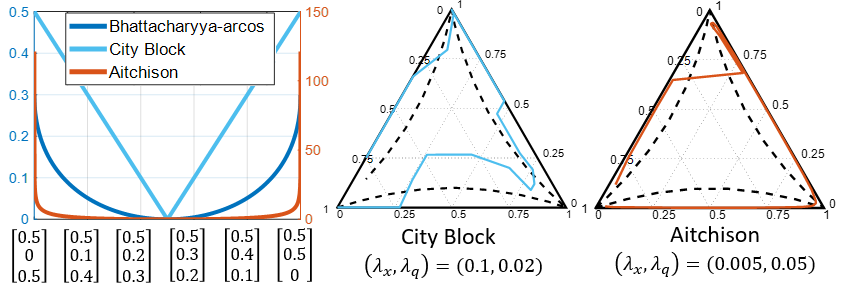}
      \caption{\rev{\textbf{Comparison of alternative simplex distances:} (left) Example of the distance profiles for various simplex distances given by $d^2(\mathbf{x}_0, \mathbf{x})$ where $\mathbf{x}_0 = [0.5, 0.25, 0.25]^\TP$ and $\mathbf{x} \in [0.5, 0, 0.5]^\TP, [0.5, 0.5, 0]^\TP$. The Bhattacharya-arccos and city-block distances are plotted against the left axis while the Aitchison, which diverges at the simplex boundary, is plotted against the right axis. (Center and right) The learned simplex trajectory using the city-block and Aitchison distance in the regularizer (Eq.~\eqref{eq:Rx}) using the same data shown in Fig.~\ref{fig:CounteractQX} which uses the Bhattacharya-arcos distance. $\lambda_x, \lambda_q$ are selected via grid search using the parameters specified in Sec.~\ref{sec:SimulatedExperiments}} and $r=200$, $n=100$.} \label{fig:SimplexDistances}
    \end{figure}
\rev{
In this work, the Bhattacharya-arccos distance was used as a regularizer on the latent state. However, a variety of simplex distances can be used \cite{martin-fernandez_measures_1998}, the choice of which may be application dependent. We compare the Bhattacharya-arccos, city-block\footnote{We approximate $\abs{x}\approx \sqrt{x^2+\epsilon}$ with $\epsilon = 1e-8$}, and Aitchison distance in Fig. \ref{fig:SimplexDistances}. The Aitchison distance diverges as one of the points moves towards the simplex boundary ($\xmb$ has one or more zeros). As a result, when using this distance as the regularizer in our DWB model, the learned simplex trajectory will also avoid edges of the simplex. This is not the case for the city-block or Bhattacharya-arccos distance which is finite for any two points on the simplex. In our simulated and real world experiments, we did not find significant differences in the ground truth error among these three simplex distances. We leave further investigation of these distance properties and their effect as regularizers to future work.
}



\ifCLASSOPTIONcaptionsoff
  \newpage
\fi

\end{document}